\pdfoutput=1

\documentclass[11pt]{article}

\usepackage[final]{acl}

\usepackage{times}
\usepackage{latexsym}
\usepackage{makecell}

\usepackage[T1]{fontenc}

\usepackage[utf8]{inputenc}

\usepackage{microtype}

\usepackage{graphicx}

\usepackage{amsmath}
\usepackage[utf8]{inputenc}
\usepackage{amsmath,amssymb,amsthm}
\usepackage{tikz}
\usepackage{algpseudocode}
\usepackage{enumerate}
\usepackage{textcomp}
\usepackage[most]{tcolorbox}
\usepackage[utf8]{inputenc} 
\usepackage{hyperref}       
\usepackage{url}            
\usepackage{booktabs}       
\usepackage{amsfonts}       
\usepackage{nicefrac}       
\usepackage{microtype}      
\usepackage{xcolor,colortbl}         
\usepackage{algorithm}
\usepackage{multirow}
\usepackage{tablefootnote}
\usepackage{xspace}
\usepackage{stmaryrd} 
\usepackage{wrapfig}   
\usepackage{paralist}
\usepackage{paralist}
\usepackage{subcaption}    
\usepackage{soul}

\sethlcolor{yellow} 

\definecolor{mygrey}{HTML}{cfcfcf}   
\definecolor{mygreen}{HTML}{8de5a1}   
\definecolor{myblue}{HTML}{a1c9f4}   
\definecolor{mypurple}{HTML}{d0bbff} 
\definecolor{myred}{rgb}{1.0, 0.6, 0.6}     
\definecolor{darkgreen}{HTML}{006401}   
\definecolor{darkred}{HTML}{8B0000}   
\definecolor{dataset}{gray}{0.9}    
\definecolor{reasoningllm}{RGB}{251,221,203}
\definecolor{ours}{RGB}{218,251,203}
\definecolor{basellm}{RGB}{222,244,252}

\newcommand{\cf}{\emph{cf.}~}

\newcommand{\cG}{\mathcal{G}}
\newcommand{\cL}{\mathcal{L}}
\newcommand{\cV}{\mathcal{V}}

\newcommand{\HB}{\mathrm{HB}}
\usepackage{pifont}
\newcommand{\cmark}{\textcolor{green}{\ding{51}}} 
\newcommand{\ymark}{\textcolor{gray}{(\ding{51})}} 
\newcommand{\xmark}{\textcolor{red}{\ding{55}}}   

\newcommand{\predicate}[1]{\textit{#1}}

\theoremstyle{definition}

\begingroup
\catcode`[=\active \catcode`]=\active

\gdef[{\llbracket} \gdef]{\rrbracket}

\def\getdelim#1#2#3#4\relax{"#4}
\global\delcode`[=\expandafter\getdelim\llbracket\relax
\global\delcode`]=\expandafter\getdelim\rrbracket\relax
\endgroup

\mathcode`\[="8000
\mathcode`\]="8000

\newcommand{\slr}{\textsc{SLR}\xspace}
\newcommand{\benchmark}{\textsc{SLR-Bench}\xspace}

\title{\slr: Automated Synthesis for Scalable Logical Reasoning}

\author{
Lukas Helff\textsuperscript{1,2,3},
Ahmad Omar\textsuperscript{1},
Felix Friedrich\textsuperscript{4}\footnotemark[2],
Antonia Wüst\textsuperscript{1},
Hikaru Shindo\textsuperscript{1},
Tim Woydt\textsuperscript{1},\\
\textbf{Rupert Mitchell\textsuperscript{1,2},
Patrick Schramowski\textsuperscript{1,2,3,5},
Wolfgang Stammer\textsuperscript{7},
Kristian Kersting\textsuperscript{1,2,3,6}}\\
$^{1}$TU Darmstadt\quad 
$^{2}$hessian.AI\quad 
$^{3}$DFKI\quad 
$^{4}$Meta FAIR\\
$^{5}$CERTAIN, Germany\quad
$^{6}$Lab1141\quad
$^{7}$MPI-Inf, SIC\\
\small{\textbf{Code:} \href{https://github.com/ml-research/ScalableLogicalReasoning}{https://github.com/ml-research/ScalableLogicalReasoning}}\\
\small{\textbf{Data:} \href{https://huggingface.co/datasets/AIML-TUDA/SLR-Bench}{https://huggingface.co/datasets/AIML-TUDA/SLR-Bench}}
}

\begin{document}
\maketitle
\bgroup
\renewcommand\thefootnote{\fnsymbol{footnote}}
\footnotetext[2]{\small work done while at TU Darmstadt/hessian.AI/Lab1141}
\egroup

\begin{abstract}

We introduce \slr, an end-to-end framework for systematic evaluation and training of Large Language Models (LLMs) via \textbf{S}calable \textbf{L}ogical \textbf{R}easoning. Given a user's task specification, \slr automatically synthesizes (i) an instruction prompt for an inductive reasoning task, (ii) a validation program, executable on model outputs to provide verifiable rewards, and (iii) the latent ground-truth rule. This process is fully automated, scalable, requires no human annotations, and offers precise control over task difficulty.
Using \slr, we create \benchmark, a benchmark comprising 19k prompts organized into 20 curriculum levels that progressively increase in relational, arithmetic, and recursive complexity. Large-scale evaluation reveals that contemporary LLMs readily produce syntactically valid rules, yet often fail at correct logical inference. Recent reasoning LLMs demonstrate improved performance but incur very high test-time computation, with costs exceeding \$300 for just 1,000 prompts.
Finally, curriculum learning via \slr doubles Llama-3-8B accuracy on \benchmark, achieving parity with Gemini-Flash-Thinking at a fraction of computational cost. Moreover, these reasoning capabilities generalize to a wide range of established benchmarks, underscoring the effectiveness of \slr for downstream reasoning.

\end{abstract}
\section{Introduction}
\begin{figure*}[!t]
    \centering
    \includegraphics[width=0.99\linewidth]{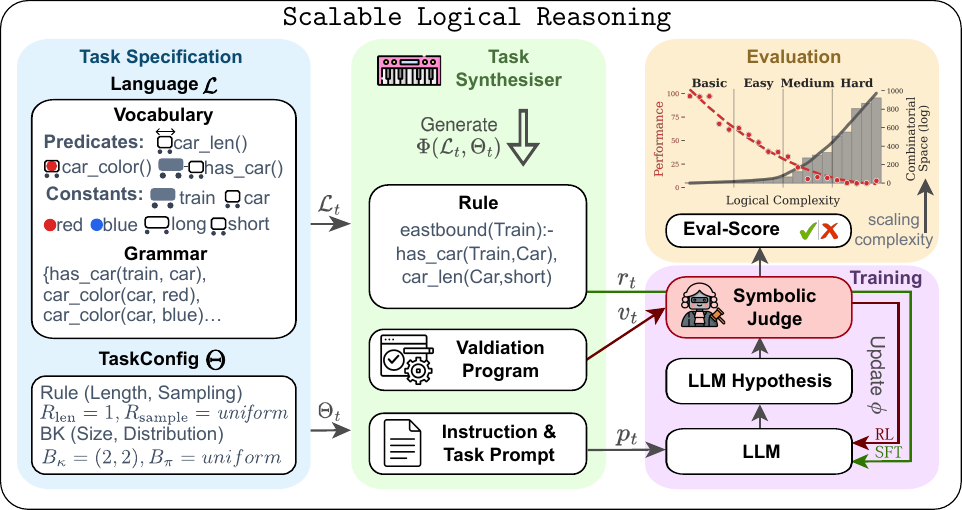}
    \caption{
Overview of the \textbf{\slr} Framework, including task specification, automated task synthesis, training, and evaluation.
\textbf{Left (blue):} Language defines vocabulary and grammar, Task Config specifies configuration parameters for the synthesis.
\textbf{Middle (green):} The task synthesizer automatically generates ground-truth rules, validation programs, and instruction prompts.
\textbf{Right (purple):} Training LLMs on logic tasks via SFT (cross-entropy) or RL (symbolic judge feedback).
\textbf{Right (orange):} Evaluates LLMs using feedback provided by the symbolic judge.
Arrows denote data and control flow through synthesis, prompting, evaluation, and downstream training loops.
}
\label{fig:metalogic-overview}
\end{figure*}
Logical reasoning is a fundamental aspect of intelligence, yet state-of-the-art AI systems still struggle with tasks that require robust reasoning and systematic generalization~\cite{delfosse2025deepreinforcementlearningagents, Kostikova2025LLLMsAD, Woydt2025FodorAP, wuest2025bongard, helff2025vlol, sinha2019clutrr}. Existing benchmarks intended to evaluate reasoning capabilities, however, primarily emphasize \textit{deductive} reasoning, where conclusions necessarily follow from given premises. This includes tasks such as math word problems~\cite{hendryckstest2021} and logic puzzles~\cite{zebralogic2025, xie2025logicrlunleashingllmreasoning, liu2025synlogic}. \textit{Inductive} reasoning, by contrast, involves inferring general rules or patterns from specific examples, which remains particularly challenging and underexplored in large language models~\cite{luo2024logiglue, xie2024memorization} (see also Tab.~\ref{tab:dataset_comparison}).

Current evaluation frameworks commonly employ constrained formats (e.g., multiple-choice) or rely on other LLMs as judges~\cite{patel2024aimeaioptimizationmultiple, Lin_ZeroEval_A_Unified_2024, lin2024wildbench}, making it difficult to assess whether models genuinely understand logical structure or are merely exploiting superficial patterns in the data. Moreover, as training sets grow, benchmark items or their paraphrases increasingly overlap with pre-training data, making apparent reasoning abilities potentially just memorization~\cite{shojaee2025illusion,xie2024memorization}.

To tackle these challenges, we introduce \textbf{\slr} (Scalable Logical Reasoning), an open-source framework for evaluating and training models in inductive logical reasoning. Given a user-defined logic task (Fig.\ \ref{fig:metalogic-overview}, left), the task synthesizer (center) automatically generates novel reasoning tasks of controllable complexity. Each task comes with (i) a latent ground-truth rule, (ii) an executable validation program, and (iii) an instruction prompt. The ground-truth rule serves as the reference answer, while the validation program deterministically evaluates any candidate hypothesis. \slr supports both systematic model evaluation (Fig.~\ref{fig:metalogic-overview}, top right) and downstream model training, via supervised finetuning or reinforcement learning with rewards provided by the integrated symbolic judge (Fig.~\ref {fig:metalogic-overview},~bottom right). \slr’s fully symbolic and automated pipeline removes the need for human annotation and prevents dataset overlap.

Leveraging \slr, we present \textbf{\benchmark} (Fig.~\ref{fig:teaser}), a ``19k task benchmark'' that forms a twenty-level curriculum of increasing logical complexity. These levels are further organized into four curriculum tiers: \emph{basic}, \emph{easy}, \emph{medium}, and \emph{hard}.
Each task is unique, with a precise and systematic assessment of inductive logical reasoning skills.
In evaluations, we find that while LLMs are generally well-versed in generating valid rules, robust logical reasoning remains challenging. Performance declines sharply as task complexity increases. Scaling model size brings only marginal improvements, while scaling test-time compute boosts reasoning, but returns diminish as complexity rises.

Beyond benchmarking, \slr enables curriculum learning, boosting reasoning both in-domain and across established reasoning benchmarks. \slr-tuned models not only surpasses conventional LLMs on \benchmark, but also outperforms reasoning LLMs, such as Gemini-2.0-flash-thinking, while using fewer inference tokens. Notably, these enhanced reasoning capabilities generalize downstream (e.g., GPQA \cite{rein2024gpqa}, and CLUTRR \cite{sinha2019clutrr}).

In sum, our contributions are:
(i) \slr, an open framework for automated synthesis and symbolic evaluation of logical reasoning in LLMs;
(ii) \benchmark, a 19k-task benchmark organized as a 20-level curriculum of increasing logical complexity, enabling both training and evaluation across a controlled reasoning spectrum;
(iii) a large-scale evaluation of LLMs on \benchmark, revealing key insights and trade-offs in model performance;
(iv) curriculum learning with \slr substantially improves both in-domain and downstream reasoning.

\section{Related Work}
\label{sec:related}
\begin{table*}[t!]
\centering
\setlength{\tabcolsep}{3pt} 
\resizebox{\textwidth}{!}{%
\begin{tabular}{@{}lccccccc@{}}
\toprule
& \textbf{Reasoning} %
& \textbf{Data} %
& \textbf{Evaluation} %
& \textbf{Task} %
& \textbf{Custom} %
& \textbf{Curriculum} %
& \textbf{Scalable} \\
\textbf{Dataset} & \textbf{Type} & \textbf{Creation} & \textbf{Methodology} & \textbf{Synthesis} & \textbf{Tasks} & \textbf{Learning} & \textbf{Complexity} \\
\midrule
LogiQA \tiny{\cite{liu2020logiqa}}           & Deduction           & Human                & MC         & \xmark & \xmark & \xmark & \xmark \\
LogiQA 2.0 \tiny{\cite{Hanmeng2023logicqa2}}       & Mix                 & Human                & MCQA/Auto         & \xmark & \xmark & \xmark & \xmark \\
FOLIO \tiny{\cite{han2022folio}}             & Deduction           & Human                & EM         & \xmark & \xmark & \xmark & \xmark \\
AbductionRules \tiny{\cite{young2022abductionrules}} & Abduction & Synthetic/Human & EM & \xmark & \xmark & \xmark & \xmark \\
FineLogic \tiny{\cite{zhou2025FineLogic}}         & Deduction           & DS collection         & Symbolic/LLM     & \xmark & \xmark & \xmark & \xmark \\
HLE \tiny{\cite{phan2025humanitysexam}} & Mix              & DS collection         & LLM               & \xmark & \xmark & \xmark & \xmark \\
Big-Bench \tiny{\cite{phan2025humanitysexam}} & Mix              & DS collection         & Mix               & \xmark & \xmark & \xmark & \xmark \\
LogiGLUE \tiny{\cite{luo2024logiglue}}          & Mix                 & DS collection         & MCQA/Auto         & \xmark & \xmark & \xmark & \xmark \\
Multi-LogiEval \tiny{\cite{patel2024multilogieval}} & Deduction                 & Synthetic             & Symbolic         & \xmark & \xmark & \xmark & \xmark \\
CLUTRR \tiny{\cite{sinha2019clutrr}} & Induction                 & Synthetic             & EM         & \xmark & \xmark & \xmark & \xmark \\
KOR-Bench \tiny{\cite{ma2024korbenchbenchmarkinglanguagemodels}} & Mix & Synthetic/Human & EM/Symbolic & \ymark & \xmark & \xmark & \xmark \\
PrOntoQA \tiny{\cite{PrOntoQA}}        & Deduction           & Synthetic                & EM         & \cmark  & \xmark & \xmark & \xmark \\
bAbI \tiny{\cite{Weston2015TowardsAQ}} & Mix                 & Synthetic             & EM         & \cmark & \xmark & \xmark & \xmark \\
SynLogic \tiny{\cite{liu2025synlogic}}      & Deduction           & Synthetic             & EM         & \cmark & \xmark & \xmark & \xmark \\
FLD \tiny{\cite{morishita202FLD}} & Deduction                 & Synthetic             & Symbolic         & \cmark & \cmark & \xmark & \ymark \\
K\&K \tiny{\cite{xie2025logicrlunleashingllmreasoning, xie2024memorization}}  & Deduction           & Synthetic                & EM         & \cmark & \xmark & \ymark & \ymark \\
ZebraLogic \tiny{\cite{zebralogic2025}}     & Deduction           & Synthetic             & EM         & \cmark & \xmark & \ymark & \ymark \\
\rowcolor{dataset} \textbf{\slr} (ours) & Induction         & Synthetic             & Symbolic         & \cmark & \cmark & \cmark & \cmark \\
\end{tabular}
}
\caption{
Comparison of \textbf{logic reasoning benchmarks.} 
\textbf{Reasoning Type}: Logical inference type (deduction, induction, abduction).
\textbf{Creation}: Dataset origin (synthetic, human-annotation, DS collection).
\textbf{Evaluation}: Output scoring (symbolic execution, multiple choice (MC), LLM, exact match (EM)).
\textbf{Task Synthesis}: Supports for tasks generation.
\textbf{Custom Tasks}: User-defined task creation (via language, grammar, or setup).
\textbf{Curriculum Learning}: Curriculum-based progression of difficulty.
\textbf{Scalable Complexity}: Supports arbitrarily scaling task complexity.
\\
(\cmark: fully supported, \xmark: not supported, \ymark: partially/limited)
}
\label{tab:dataset_comparison}
\end{table*}
\noindent \textbf{Evaluating LLMs' Logical Reasoning.} Tab.~\ref{tab:dataset_comparison} provides an overview of existing logical reasoning benchmarks in terms of inference types, dataset origins, and evaluation formats. Notable datasets include LogiQA/2.0~\cite{liu2020logiqa, Hanmeng2023logicqa2}, FOLIO~\cite{han2022folio} (deductive reasoning), AbductionRules~\cite{young2022abductionrules} (abductive reasoning), bAbI~\cite{Weston2015TowardsAQ}, and CLUTRR~\cite{sinha2019clutrr} (synthetic QA with inductive reasoning). Aggregate testbeds such as BIG-Bench~\cite{kazemi2025bigbenchextrahard, suzgun2023bigbench}, HLE~\cite{phan2025humanitysexam}, FineLogic~\cite{zhou2025FineLogic}, and LogiGLUE~\cite{luo2024logiglue} span a range of tasks and inference styles. Other benchmarks, such as Proofwriter, PrOntoQA, FLD, Multi-LogiEval, SynLogic, ZebraLogic, and the K\&K Sandbox~\cite{tafjord2021proofwriter, PrOntoQA, morishita202FLD, patel2024multilogieval, liu2025synlogic, zebralogic2025, xie2025logicrlunleashingllmreasoning, xie2024memorization} generate tasks from fixed ontologies, often with exact-match evaluation. Classic ILP datasets (e.g. Mutagenesis~\cite{debnath1991mutagenesis}) lack scalability and natural-language integration. In contrast, \slr introduces scalable, curriculum-based synthesis with controllable difficulty and verifiable evaluation, addressing key gaps in prior works.

\noindent \textbf{Limits and Promises of Reasoning LLMs.} LLMs like GPT-4~\cite{openai2023gpt4}, Llama-3~\cite{grattafiori2024llama3herdmodels}, and Qwen~\cite{bai2023qwen} can handle basic reasoning and coding tasks but often struggle with true abstraction~\cite{shojaee2025illusion, xie2024memorization}. Recent \emph{reasoning LLMs} attempt to bridge this gap by scaling \emph{test-time} compute. Systems like OpenAI’s \textit{o1}/\textit{o3}~\cite{openai_o3_system_card_2025} or DeepSeek-R1~\cite{DeepSeekAI2025DeepSeekR1IR} generate and re-rank thousands of reasoning traces per query, achieving state-of-the-art results on, e.g., math or coding~\cite{quan2025codeelo, hendryckstest2021, rein2024gpqa, eval-harness}. 
However, these gains come at a steep cost~\cite{fan2025cothinktokenefficientreasoninginstruct, kim2025costdynamicreasoningdemystifying}. Some studies question whether such models truly learn logical structure or merely exploit surface-level patterns~\cite{fan2025cothinktokenefficientreasoninginstruct, shojaee2025illusion,xie2024memorization}. Curriculum learning has been shown to enhance training robustness and generalization~\cite{bengio2009curriculum, bursztyn2022compositional}, yet no prior framework offers a flexible framework for task synthesis for automatic curriculum generation with symbolic evaluation for reasoning at scale. \slr addresses this gap.

\section{\slr: Automatic Benchmark Synthesis} \label{sec:metalogic}
\slr is a scalable methodology for systematically generating, evaluating, and training LLMs on inductive reasoning tasks. Its goal is to automate the creation of diverse, challenging logical reasoning benchmarks, embedded as natural language prompts, with model outputs that can be efficiently verified via symbolic execution of Inductive Logic Programming (ILP) programs~\cite{ILP_Muggleton,ILP_Cropper}. The overall pipeline (Fig.~\ref{fig:metalogic-overview}) has three main stages: \textit{task specification}, \textit{synthesis}, and \textit{evaluation/training}.

\noindent \textbf{Task Notation.}\label{sec:ilp-task} \slr follows the Learning-from-Entailment (LFE) paradigm~\citep{Raedt1997LogicalSF} in ILP, where each task is defined as $\mathcal{I} = (B, E^+, E^-)$ consisting of background knowledge $B$, positive examples $E^+$, and negative examples $E^-$. A candidate hypothesis $H$ \emph{solves} the task if and only if $B \cup H \models E^+$ and $B \cup H \not\models E^-$. Logical entailment ($\models$) is evaluated via Prolog execution under the closed-world assumption (details see App.~\ref{app:sec:fol}).

\noindent \textbf{Motivating Example.}
Consider a simple train domain where each train consists of cars described by attributes like color, length, or roof type.
The learning goal is to induce a rule, e.g., ``the train has a yellow car.''
Here, the \textit{background knowledge}~($B$) contains all known facts about each train, e.g., which cars it includes and their attributes.
The \textit{positive examples}~($E^+$) all have yellow cars, while the \textit{negative examples}~($E^-$) do not have any yellow cars.
During synthesis, \slr automatically generates such tasks by varying both the background knowledge and the ground-truth rule, effectively scaling the complexity of the reasoning problem.
\begin{algorithm}[t]
\footnotesize
\caption{Task Synthesizer}
\label{alg:synth}
\begin{algorithmic}[1]
\Require $\mathcal{L},\; B_\pi,\; \kappa_{\text{pos}},\; \kappa_{\text{neg}},\; R_{\text{sample}}, R_{\text{len}}$
\State $B\gets\varnothing$, $E^{+}\gets\varnothing$, $E^{-}\gets\varnothing$
\Statex{\# Rule Synthesis}
\State $R^\star \gets \textsc{RuleGenerator}(\mathcal{L},R_{len},R_{sample})$ 
\While{$|E^{+}|<\kappa_{\text{pos}}$ \textbf{or} $|E^{-}|<\kappa_{\text{neg}}$}
    \Statex{\# Background Synthesis}
    \State $b \gets \textsc{BackgroundGenerator}(\mathcal{L}, B_\pi)$
    \State $(y,q) \gets \textsc{AssignLabel}(R^\star,b)$
    \Statex{\# Stratified Rejection Sampling}
    \If{$y=1$ \textbf{and} $|E^{+}|<\kappa_{\text{pos}}$} \Comment{accept positive}
        \State $B\gets B\cup \{b\}$;\; $E^{+}\gets E^{+}\cup\{q\}$ 
    \ElsIf{$y=0$ \textbf{and} $|E^{-}|<\kappa_{\text{neg}}$} \Comment{accept negative}
        \State $B\gets B\cup \{b\}$;\; $E^{-}\gets E^{-}\cup\{q\}$
    \Else                                    
        \State \textbf{continue}\Comment{reject sample}
    \EndIf
\EndWhile
\State program $\gets \textsc{ValidationProgram}(B,E^{+},E^{-})$
\State prompt $\gets \textsc{PromptGenerator}(B,E^{+},E^{-})$
\State \Return ($R^\star$, program, prompt)
\end{algorithmic}
\end{algorithm}
\subsection{Task Specification (Input)}
The \slr synthesizer is controlled by the Task Language $\mathcal{L}$, which defines the logical vocabulary and grammar, and the Task Configuration~$\Theta$, which controls the generation process (see Fig.~\ref{fig:metalogic-overview}, left).

\noindent \textbf{Language Specification ($\cL$):}
We define a language $\cL = (\cV, \cG)$ that specifies the building blocks for task generation. 
The Vocabulary $\mathcal{V}$ comprises a set of constant, function, and predicate symbols that form the syntax for generating rules, examples, and background knowledge. The vocabulary induces the Herbrand base $\HB(\mathcal{V})$, which is the set of all syntactically valid ground atoms (facts)~\cite{lloyd2012foundations}. 
The Grammar $\mathcal{G}$ is formed by a set of semantic rules that filter the Herbrand base to include only meaningful atoms, $\HB_\mathcal{G}(\mathcal{V})$. 
For instance, a color can be assigned to a car (\text{car\_color(car, red)}) but not to semantically incompatible objects.

\noindent \textbf{Task Configuration ($\Theta$):}
The configuration parameters $\Theta = \langle R_{\text{sample}}, R_{\text{len}}, B_{\pi}, \kappa \rangle$ give control over the synthesis process. The (i) Rule Sampling Policy ($R_{\text{sample}})$ controls the synthesis of the ground truth rule $R^{\star}$, which can either be sampled randomly (Uniform Sampling) or generated via an LLM (LLM-Guided Generation). To ensure the LLM produces diverse and challenging logic rules, we leverage an exhaustive prompt (see App.~\ref{sec:llmprompt}) that covers a wide array of logical structures and Prolog features (containing arithmetics, recursions, variables, cuts, or comparison operators, etc.).
The (ii) Rule Length ($R_{\text{len}}$) specifies the number of literals in the body of the ground-truth rule $R^\star$. Structural predicates such as has\_car(T,C), which merely link entities and impose no constraints, are not counted.
The (iii) Background Sampling Policy ($B_{\pi}$) defines a probability mass function that assigns a selection probability to each ground atom in the grammar-filtered Herbrand base $\HB_\mathcal{G}(\mathcal{V})$, enabling designers to encode priors on the data distribution (e.g., \emph{uniform}). We further introduce mirror sampling, where backgrounds for $(E^+, E^-)$ are identical except for atoms relevant to $R^\star$. For each positive example, a corresponding “mirror” negative is created by randomly altering the rule-specific atoms.
The (vi) Problem Size ($\kappa = (\kappa_{\text{pos}}, \kappa_{\text{neg}})$) specifies the target number of positive ($|E^+|$) and negative ($|E^-|$) examples. This directly controls the size and class balance of each generated task.
\begin{figure}
    \centering
    \includegraphics[width=.99\linewidth]{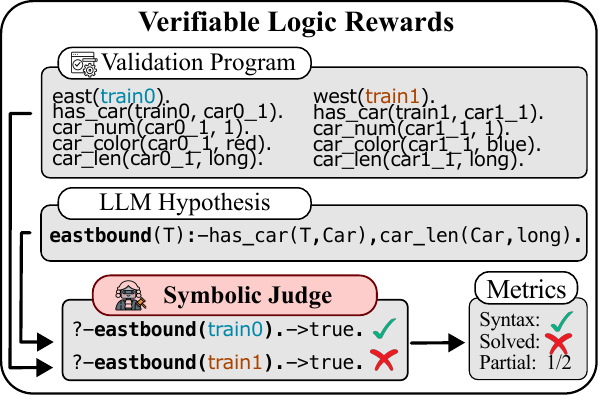}
    \caption{\textbf{Verifiable Logic Rewards:} A candidate hypothesis is evaluated by executing it against the validation program. It outputs three metrics: syntactic validity (binary), perfect task completion (binary), and a partial score for the fraction of correctly classified examples.}
    \label{fig:judge}
\end{figure}
\begin{figure}
  \centering
  \includegraphics[width=.99\linewidth]{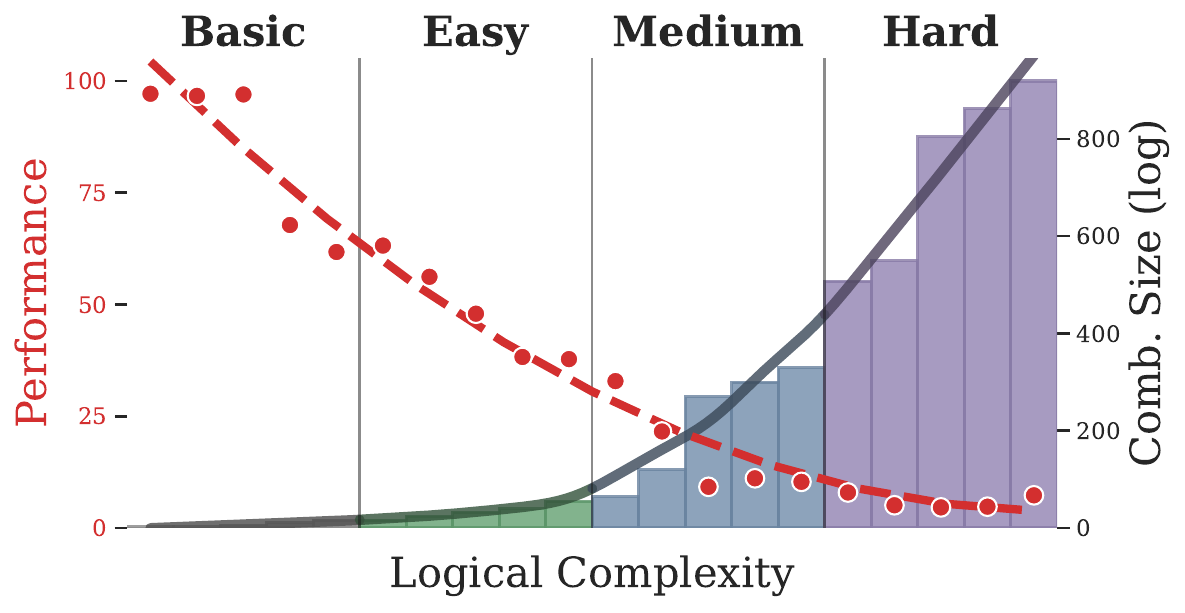}
  \caption{Overview of \textbf{\benchmark}: The benchmark curriculum spans from basic to hard tasks with increasing logical and combinatorial complexity (bars, right y-axis). As logical complexity increases, Model performance (red, left y-axis) declines, highlighting current LLMs' limitations on more challenging reasoning tasks.}
  \label{fig:teaser}
\end{figure}
\subsection{Task Synthesis (Generation)}
The task synthesizer (Fig.~\ref{fig:metalogic-overview}, center) is an automated process detailed in Alg.~\ref{alg:synth}. Given a task specification, the synthesizer generates complete and solvable ILP problems. It comprises two main phases: rule synthesis and background synthesis.

\noindent \textbf{Rule Synthesis (Alg.~\ref{alg:synth}, line 2).}
The process begins with the \textsc{RuleGenerator} creating a latent, ground-truth rule $R^\star$. This rule represents the underlying logical pattern that a model is expected to induce. The generation is guided by pre-defined parameters ($R_{\text{len}}$, $R_{\text{sample}}$) that control the length and generation policy for the rule. The resulting rule is a syntactically valid definite clause of the form $h \texttt{:-} b_1,~\dots,~b_{R_{\text{len}}}$.

\noindent \textbf{Background Synthesis (Alg.~\ref{alg:synth}, lines 3-13).}
Once $R^\star$ is fixed, the synthesizer enters a loop to construct the background knowledge $B$ and the label sets $E^+$ and $E^-$. This loop executes three steps until the desired number of positive and negative examples is generated:

(i) \textbf{Sample Background:} The \textsc{BackgroundGenerator} samples a set of ground atoms specifying the properties and relationships of the background instance. Ground atoms are drawn from the probability mass function $B_\pi$ over $\HB_\mathcal{G}(\mathcal{V})$.

(ii) \textbf{Assign Label:} The function determines whether a query atom $q$ (that is the ground atom of the target predicate $h$) is logically entailed by the sampled background $b$ and the ground-truth rule $R^\star$ (i.e., whether $b \cup R^\star \models q$ holds). This produces a label (positive or negative) for the query. We denote the labeling function as: $\textsc{AssignLabel}(R^\star, b) = (1, q)$ if $b \cup R^\star \models q$, and negative otherwise, $(0, q)$.

(iii) \textbf{Accept/Reject Sample:} To ensure the desired class balance, a stratified rejection sampling strategy is used to populate the example sets. The generated background $b$ and query $q$ are accepted only if the corresponding example set ($E^+$ or $E^-$) is not yet full, as specified by the task size parameter ($\kappa$). If accepted, $b$ is added to the task's main background knowledge $B$, and $q$ is added to the appropriate example set. Otherwise, it is discarded.

\begin{figure*}[t!]
    \centering
    \includegraphics[width=.99\linewidth]{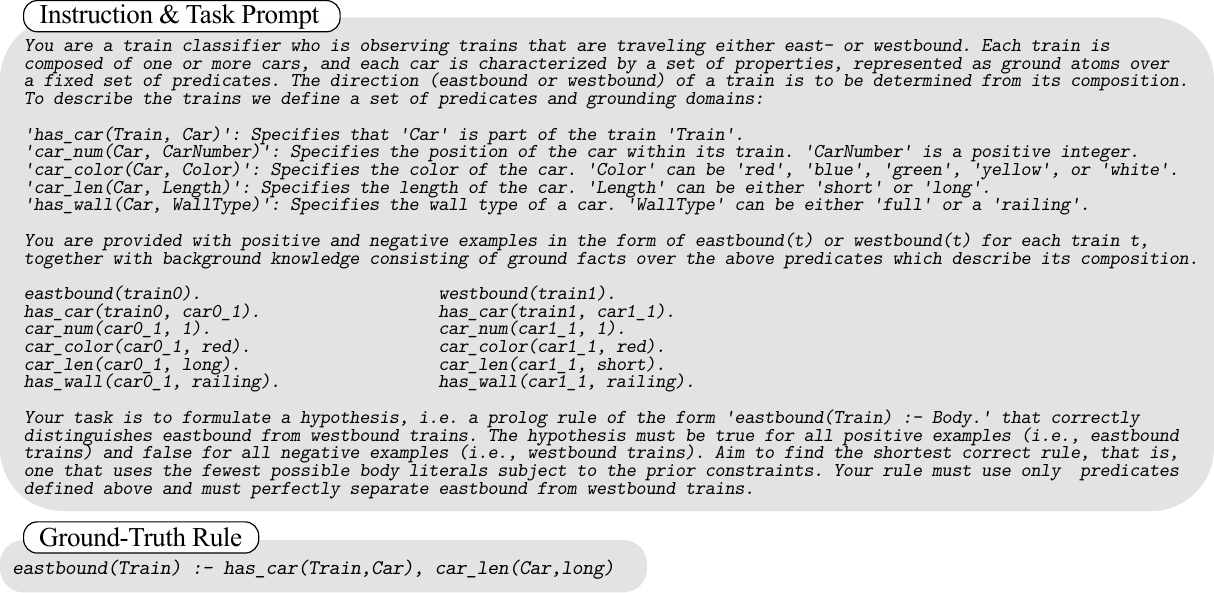}
    \caption{Illustrative \textbf{prompt} and \textbf{ground-truth rule} generated by \slr{} (Level 1, \benchmark). Language ($\mathcal{L}$): 5 predicates, 1 car variable per train. Task configuration ($\Theta$): $\kappa = (1,1)$ (one positive and one negative example); $B_\pi =$ mirror; $R_{\text{len}} = 1$; $R_{\text{sample}} =$ uniform. The prompt provides the full ILP instance, including background $B$, positive/negative examples ($E^+, E^-$), and natural-language instructions for the learning task.}
    \label{fig:prompt}
\end{figure*}
\noindent \textbf{Synthesizer Outputs (Alg.~\ref{alg:synth}, lines 14–17).}
For each task, the synthesizer generates three outputs: (1) the \textit{latent ground-truth rule} $R^\star$; (2) a \textit{validation program}, an executable logic program encoding $(B, E^+, E^-)$ for automatic evaluation; and (3) an \textit{instruction prompt} presenting the task in natural language or Prolog, ready for LLM input. See App.~Fig.~\ref{app:fig:synthesis} for an example synthesis run.

\subsection{Training and Evaluation}
The final stage, shown in Fig.~\ref{fig:metalogic-overview} (right), uses the synthesized task to evaluate and train models.

\noindent \textbf{\textsc{Verifiable Logic Rewards}.}
In \slr verifiable rewards are implemented via the Symbolic judge  (see Fig.~\ref{fig:judge}), which is used for both training and evaluation. It deterministically assesses candidate hypotheses for logic tasks by executing them against a validation program and providing rewards signals. Specifically, it checks whether all positive examples ($E^+$) are entailed and all negative examples ($E^-$) are not entailed (see App.~\ref{app:sec:logicrewards}).

\noindent \textbf{Model Evaluation.}
\slr streamlines the creation of logical reasoning benchmark datasets for systematic model evaluation. By specifying various combinations of task language and configuration, users can automatically generate diverse tasks that span a broad range of domains, prompt styles, and reasoning complexities. Each synthesized task comprises a natural language prompt, a ground-truth rule, and an executable validation program.

\noindent \textbf{Model Training.}
\slr enables automated training loops, with two types of model feedback. In supervised fine-tuning (SFT), the ground-truth rule $R^\star$ acts as the training target, enabling gradient-based updates from predicted rules. In reinforcement learning, the verifiable logic rewards provide a reward signal (RLVR) that guides policy updates. This cohesive pipeline enables scalable generation, evaluation, and training of LLMs, driving systematic advances in logical reasoning capabilities.

\noindent \textbf{Novelty and Systematic Generalization.}
The generator enforces a strict separation between training and test splits, ensuring that ground-truth rules ($R^\star$) and background knowledge ($B$) are entirely distinct. As \slr is fully synthetic, overlap with existing data is statistically negligible, making it robust against data leakage and memorization. This enables assessing whether the model is capable of systematically generalizing to entirely new rules.
\section{\benchmark: Instantiating \slr} \label{sec:metabench}

With \benchmark, we instantiate \slr as a 20-level curriculum of logical reasoning tasks with increasing complexity (see Fig.~\ref{fig:teaser}). Each level specifies its own language $\mathcal{L}$ and configuration $\theta$, producing a total of 19k generated reasoning tasks. Each level contains 1k train\footnote{The trainsets for levels 1-3 are smaller (26, 234, and 793)}, 10 eval, and 50 test samples. Each task comes with (i) a generated latent ground-truth rule, (ii) the corresponding validation program, and associated instruction prompt for the task. An illustrative example for prompts and ground-truth rules can be found in Fig.~\ref{fig:prompt}.

\noindent \textbf{Design rationale.}
The logic task is inspired by the V-LoL \textit{trains} domain~\cite{helff2025vlol, michalski1980, Mitchell1997MachineLI}, chosen for three main reasons. First, its hierarchical structure (trains $\rightarrow$ cars $\rightarrow$ attributes) naturally yields first-order rules richer than simple lookups yet more tractable than full theorem proving. Second, its small, discrete attribute domains enable precise control over task complexity. Third, it offers a clean proof of concept for scalable logic synthesis, being fully synthetic and easily extendable to new domains.

\begin{table*}[t]
    \centering
    \resizebox{.99\linewidth}{!}{
    \begin{tabular}{llllllllll@{}}
        \multicolumn{7}{c}{\textit{Curriculum Learning (\benchmark{})}}\\
         & LRL \scriptsize{($\uparrow$0-20)}& Syntax\scriptsize{($\uparrow$\%)} & Basic\scriptsize{($\uparrow$\%)} & Easy\scriptsize{($\uparrow$\%)} & Medium\scriptsize{($\uparrow$\%)} & Hard\scriptsize{($\uparrow$\%)} \\
        \midrule
        Llama3.1-8b-it & 3.8 & \phantom{0}80 & 70 & \phantom{0}7 & 0 & 0  \\
        \rowcolor{gray!15}\textbf{Llama3.1-8b-it-SLR}
        & \textbf{8.7 }\textcolor{darkgreen}{\small{(+4.9)}}
        & \textbf{100 }\textcolor{darkgreen}{\small{(+20)}}
        & \textbf{96 }\textcolor{darkgreen}{\small{(+26)}}
        & \textbf{54 }\textcolor{darkgreen}{\small{(+47)}}
        & \textbf{18 }\textcolor{darkgreen}{\small{(+18)}}
        & \textbf{5 }\textcolor{darkgreen}{\small{(+5)}} \\  
        \multicolumn{7}{c}{\textit{Downstream Reasoning Performance}}\\
        & SLR-Homes {\scriptsize($\uparrow$\%)}& MMLU {\scriptsize($\uparrow$\%)}& MMLU-Stats {\scriptsize($\uparrow$\%)}& CLUTRR {\scriptsize($\uparrow$\%)}& LogiQA {\scriptsize($\uparrow$\%)}& LogiQA2 {\scriptsize($\uparrow$\%)}\\
        \midrule
        Llama3.1-8b-it & 10.3 & 63.3 & 42.6 & 25.5 & 30.1 & 34.3 \\
        \rowcolor{gray!15}\textbf{Llama3.1-8b-it-SLR}
        & \textbf{42.3 }\textcolor{darkgreen}{\small{(+32)}}
        & \textbf{66.1 }\textcolor{darkgreen}{\small{(+2.8)}}
        & \textbf{59.7 }\textcolor{darkgreen}{\small{(+17)}}
        & \textbf{32.1 }\textcolor{darkgreen}{\small{(+6.6)}}
        & \textbf{31.0 }\textcolor{darkgreen}{\small{(+0.9)}}
        & \textbf{39.4 }\textcolor{darkgreen}{\small{(+5.2)}} \\
        & GPQA {\scriptsize($\uparrow$\%)}& GPQA-Ext. {\scriptsize($\uparrow$\%)}& GPQA-Dia. {\scriptsize($\uparrow$\%)}& ARC-Easy {\scriptsize($\uparrow$\%)}& ARC {\scriptsize($\uparrow$\%)}& HellaSwag {\scriptsize($\uparrow$\%)}\\
        \midrule
        Llama3.1-8b-it
        & 31.7 & 26.9 & 21.7 & 81.4 & 52.7 & 57.4 \\
        \rowcolor{gray!15}\textbf{Llama3.1-8b-it-SLR}
        & \textbf{32.8 }\textcolor{darkgreen}{\small{(+1.1)}}
        & \textbf{33.0 }\textcolor{darkgreen}{\small{(+6.1)}}
        & \textbf{28.3 }\textcolor{darkgreen}{\small{(+6.6)}}
        & \textbf{82.8 }\textcolor{darkgreen}{\small{(+1.4)}}
        & \textbf{54.6 }\textcolor{darkgreen}{\small{(+1.9)}}
        & \textbf{58.9 }\textcolor{darkgreen}{\small{(+1.5)}} \\
    \end{tabular}
        }
    \caption{
        \textbf{Curriculum Learning and Generalization.}
        Benchmark scores {\scriptsize($\uparrow$)} for base and SLR-tuned models on \benchmark{} and downstream benchmarks; LRL measuring cumulative curriculum progress. The tuned model surpasses the baseline across all curriculum stages, while generalizing to other downstream reasoning tasks.
     }
    \label{tab:downstream}
\end{table*}
\noindent \textbf{Languages.}
Each curriculum level is parameterized by level-specific language $\mathcal{L}$, detailed in App.~\ref{sec:app:grammar}. The vocabulary includes mutually exclusive class labels \predicate{eastbound} and \predicate{westbound}, which serve as the targets for classification tasks $(E^+,E^-)$. Background knowledge $B$ is sampled from a vocabulary of five predicates (\predicate{has\_car}, \predicate{car\_num}, \predicate{car\_color}, \predicate{car\_len}, and \predicate{has\_wall}) with their respective grounding domains defined in App.~\ref{sec:app:grammar}. As curriculum levels increase, the vocabulary expands monotonically by introducing new predicates and grounding domains selected from a predefined set. Semantic coherence is ensured by constraining predicate groundings (e.g., only colors as arguments for \predicate{car\_color}), and by enforcing mutually exclusive constraints across predicates (e.g., passenger cars cannot carry payloads).

\noindent \textbf{Task configs.}
Each curriculum level is parameterized by level-specific settings of $\theta$, summarized in App.~Tab.~\ref{app:tab:task_spec} and supplied directly to the synthesizer (Alg.~\ref{alg:synth}). Problem size ($\kappa$) increases steadily across levels, maintaining an equal balance of positive and negative samples.
Levels 1–5 use a \textit{mirror} sampling policy for background knowledge, generating simple, nearly identical east- and westbound trains that differ only in ground atoms relevant to $R^\star$. From level 6 onward, the background is sampled uniformly from the filtered Herbrand base, increasing diversity. Rule generation is uniform for the basic levels; from level 6, 30\% of rules are LLM-guided, introducing greater variety in variables, arithmetic, recursion, and more.

\noindent \textbf{SLR-Curriculum.}
\benchmark\ comprises 20 levels across four tiers: \textit{basic}, \textit{easy}, \textit{medium}, and \textit{hard}. Each level systematically increases complexity by expanding task size ($\kappa$), adding new car constants and predicates, lengthening rules, and varying both the background knowledge and rule sampling policy; see App.~Sec.~\ref{app:sec:task_spec}, Tab.~\ref{app:tab:task_spec}. As a result, the combinatorial space of possible tasks grows exponentially, and later levels become progressively harder and require deeper reasoning beyond surface cues.

\noindent \textbf{SLR-Homes.}
Alongside the main curriculum, we introduce an OOD dataset that instantiates SLR in a new symbolic domain. It uses a distinct vocabulary, relational structure, and target concept to define house classification tasks based on attributes. SLR-Homes enables the assessment of cross-domain generalization and comprises 500 tasks across five difficulty levels, increasing in complexity.

\noindent \textbf{Intended Use.}
\benchmark\ is designed for two complementary purposes.
(1) As a \emph{static} benchmark, it enables fine-grained evaluation of an LLM’s reasoning abilities across tasks of increasing logical complexity. It is also easily extensible to accommodate future improvements in model capabilities.
(2) As a \emph{dynamic} curriculum, it serves as a training backbone, supplying structured reasoning tasks and feedback to enhance reasoning in both conventional and reasoning LLMs.

\begin{table*}[th]
\centering
\begin{tabular}{lrrrrrrrr@{}}
\toprule
& LRL & Syntax
& \multicolumn{4}{c}{Logical-Reasoning Acc. \scriptsize{(\%)$\uparrow$}} 
& \multicolumn{2}{c}{Total Compute} \\
\cmidrule(lr){4-7}
\cmidrule(lr){8-9}
Model & \scriptsize{($\uparrow$0-20)}
& Score \scriptsize{($\uparrow$\%)}
& Basic & Easy & Medium & Hard & Tokens \scriptsize{($\downarrow$M)} & Costs \scriptsize{($\downarrow$\$)}\\
\midrule
\cellcolor{reasoningllm!70}o3                & \textbf{15.5} &  80 & {99} & \textbf{93} & \textbf{74} & 45 & 4.30 &207.24 \\
\cellcolor{reasoningllm!70}gpt-5            & 15.4 & 98 & \textbf{100} & 90 & 72 & \textbf{46} & 16.40 & 103.13 \\
\cellcolor{reasoningllm!70}gpt-5-mini-high & 14.2 & 0.94 & 99  & 83.0 & 63.0 & 38.0 & 13.12 & 27.98 \\
\cellcolor{reasoningllm!70}gpt-5-mini      & 12.0 & 95 & 99  & 82 & 41 & 20 & 4.90  & 11.54 \\
\cellcolor{reasoningllm!70}o1                & 11.9 &  68 & 92 & 89 & 41 & 15 & 5.19 &364.72 \\
\cellcolor{reasoningllm!70}gpt-5-mini-low  & 9.7  & 0.88 & 97  & 71.0 & 20.0 & 7.0  & 1.17  & 4.07 \\
\cellcolor{reasoningllm!70}R1-Llama-70B\footnotemark[2]   &  8.8 & 75 & 98 & 67 &  8 &  4 & 11.61 &5.33 \\
\cellcolor{reasoningllm!70}gpt-5-nano      & 8.5  & 99 & 97  & 61 & 10 & 2  & 6.17  & 2.81 \\
\cellcolor{reasoningllm!70}Gemini-thinking\footnotemark[1]&  8.6 & 83 & 93 & 65 & 13 &  1 & ---\phantom{0}   & ---\phantom{0}    \\
\cellcolor{basellm!70}gpt-4.5-prev           & 7.3 & \textbf{100} & 94 & 47 &  5 &  1 & 0.37 &576.40 \\
\cellcolor{basellm!70}gpt-4o                 & 6.2 & \textbf{100} & 93 & 29 &  2 &  0 & 0.26 &20.03 \\
\cellcolor{basellm!70}Llama 3.3-70B          & 5.6 & \textbf{100} & 90 & 22 &  1 &  0 & 0.49 & 0.82 \\
\cellcolor{basellm!70}gpt-4-turbo            & 5.4 & \textbf{100} & 89 & 18 &  2 &  0 & 0.41 &81.30 \\
\cellcolor{basellm!70}Llama 3.1-8B           & 3.8 & 80  & 70 &  7 &  0 &  0 & 0.20 & 0.14 \\
\end{tabular}
\begin{minipage}{\textwidth}
\footnotesize \hspace{4em}
$^1$Gemini-2.0-flash-thinking-exp-01-21  \phantom{0}$^2$DeepSeek-R1-Distill-Llama-70B \phantom{00}--- information not available
\end{minipage}
\caption{\textbf{\benchmark Leaderboard}. We report the models' Logical Reasoning Level (LRL), syntax score, stage-specific logical reasoning accuracy (basic, easy, medium, hard), total completion tokens, and inference cost. Higher LRL and accuracy indicate superior logical reasoning; lower compute, greater efficiency. Performance drops as complexity increases, while Reasoning LLMs (orange) consistently outperform conventional LLMs (blue). \label{tab:logic-benchmark}}
\end{table*}
\section{LLMs Can’t Do Induction at Scale}\label{sec:experiments}
We benchmark and train LLMs on \benchmark, assessing reasoning, syntactic correctness, and computational efficiency across the four difficulty tiers: \textit{basic}, \textit{easy}, \textit{medium}, and \textit{hard}. Our analysis highlights key trends, common failure modes, and the effectiveness of curriculum-based logic-tuning.

\noindent \textbf{Training Setup.} We investigate how LLMs benefit from curriculum training on \benchmark with SFT (for more details see  App.~Sec.~\ref{sec:training_details}). To prevent data leakage, we ensure that no prompts or rules from the test set are included in the training set. 

\noindent \textbf{Evaluation Setup.} All models are evaluated in a zero-shot setting using \benchmark prompts, with a single attempt per task (pass@1). We report the following metrics:
(i)~\textit{Logical Reasoning Level (LRL):} A cumulative score over all curriculum levels $L$, where $\#\text{solved}_{\ell}$ and $\#\text{tasks}_{\ell}$ denote the number of solved and total tasks at level $\ell$:
$\text{LRL}=\sum_{\ell=1}^{L}\frac{\#\text{solved}_{\ell}}{\#\text{tasks}_{\ell}}$
(ii)~\textit{Syntax Score:} The proportion of predicted logic rules that are syntactically valid.
(iii)~\textit{Logical-Reasoning Accuracy:} The fraction of correct solutions per complexity tier.
(iv)~\textit{Compute:} The aggregate completion tokens and computational cost for each the models. 
For further details on downstream evaluations, see App.~Sec.~\ref{sec:training_details}, and for pricing, refer to App.~Tab.~\ref{tab:pricing}.
\subsection{Analysis and Key Findings}
In this section, we analyze the key effects of \slr training. We first report downstream reasoning gains for models trained with \slr, namely Llama-3.1-8B-IT (Tab.~\ref{tab:downstream}) and Qwen3-8B (App. Tab.~\ref{app:tab:downstream_ext}). We then benchmark SOTA conventional and reasoning LLMs on \benchmark (Tab.~\ref{tab:logic-benchmark}), with an extended leaderboard provided in App. Tab.~\ref{tab:extended-leaderboard}.

\noindent \textbf{\slr Boosts Downstream Reasoning.}
Curriculum learning on \benchmark yields substantial gains in both in-domain and downstream reasoning (Tab.~\ref{tab:downstream}). \slr-tuned models consistently outperform conventional LLMs on \benchmark (\cf Tab.~\ref{tab:logic-benchmark}) and even surpass dedicated reasoning models such as Gemini-2.0-flash-thinking, while requiring significantly fewer inference tokens and lower compute. Performance on SLR-Homes further demonstrates robust zero-shot transfer to previously unseen domains, suggesting that \slr learns domain-agnostic reasoning capabilities rather than dataset-specific heuristics. Beyond \benchmark, these gains transfer to a broad set of established reasoning benchmarks, including MMLU High School Statistics (+17) \cite{hendryckstest2021}, CLUTRR \cite{sinha2019clutrr}, LogicQA \cite{liu2020logiqa}, LogicQA2 \cite{Hanmeng2023logicqa2}, ARC \cite{Clark2018ThinkYH}, HellaSwag \cite{zellers2019hellaswag}, and GPQA variants \cite{rein2024gpqa}. We hypothesize that \slr training introduces an inductive bias toward systematic rule induction over structured relational contexts. Models are encouraged to internalize abstractions for identifying relational structure, composing and applying reasoning patterns. These abstractions generalize to downstream deductive and knowledge-intensive tasks, where performance likewise depends on structured reasoning.

\noindent \textbf{Curriculum Order Matters.}
An ablation study (App.~\ref{app:sec:curriculum}) examines how curriculum order affects training performance by comparing ordered, random, and reverse progressions. Training with the ordered curriculum achieves the best results, confirming that structured progression supports smoother and more stable learning.

\noindent \textbf{Curriculum Levels Modulate Task Complexity: LLMs Break Down as Complexity Increases.}
\benchmark creates a controlled gradient in logical complexity as model performance steadily declines throughout the curriculum levels (\cf Fig.~\ref{fig:teaser}, Tab.~\ref{tab:logic-benchmark}). Most models readily solve \textit{basic} levels. Base LLMs already falter on \textit{easy} ones, solving fewer than half. Reasoning LLMs perform better, yet their accuracy drops sharply at the \textit{medium} tier, and none succeed on \textit{hard}. This pattern is also reflected in the LRL score, empirically indicating how far each model can progress before performance collapses. An ablation study (App.~\ref{app:sec:synthesis_parameters}) further shows how increasing rule length or problem size, and switching from “uniform” to “LLM-guided” or “mirror” to “uniform” sampling, influences task complexity, confirming that \slr provides fine-grained control over logical complexity.

\noindent \textbf{Test-time Scaling Improves Reasoning, but Returns Diminish and Costs Escalate.} Reasoning LLMs clearly outperform the base models; not even the best base model is able to match any of the reasoning LLMs (\cf Tab.~\ref{tab:logic-benchmark}). CoT prompting provides modest gains ($\approx5\%$, App.~\ref{app:sec:cot}). Further, scaling test-time computation comes at a steep cost as moving from GPT-4o to \textit{o3} doubles accuracy, but considerably increases the number of completion tokens (1777\%) and thus the computational costs (1034\%). Moreover, scaling test-time compute on the same model also boosts overall performance (see App.~\ref{app:sec:scaling_compute}), but does not guarantee higher accuracy across all tasks. E.g., while \textit{o4-mini-high} typically outperforms \textit{o4-mini} (LRL: 12.8 vs.~12.3), it underperforms on the medium tier (40\% vs.~52\%). This plateau effect demonstrates that, beyond a certain threshold, additional compute may yield diminishing or even negative returns.

\noindent \textbf{Scaling Model Parameters Brings Limited Gains in Logical Reasoning.} 
Increasing model size yields only marginal gains in logical reasoning. While larger models like Llama-3-70B and GPT-4.5-prev generally outperform their smaller counterparts, returns diminish as improvements are increasingly modest (see Tab.~\ref{tab:logic-benchmark}). As even the largest base models still fall short of the reasoning models, a capability gap remains that suggests that scaling model parameters alone does not guarantee substantial advances in logical reasoning capabilities.

\noindent \textbf{Common Error Modes.} 
Base LLMs reliably produce syntactically valid rules, as reflected by their high syntax scores (Tab.~\ref{tab:logic-benchmark}). Reasoning models achieve slightly lower syntax scores, particularly on tasks of higher complexity.
However, the dominant performance bottleneck is semantic rather than syntactic, as evidenced by the persistent gap between syntax and LRL. An error analysis (App.~\ref{app:sec:error_modes}) identifies five recurring failure modes: (i) over-generalized rules, (ii) under-generalized rules, (iii) failure of abstraction, and (iv) degenerate reasoning loops. Further, accuracy decreases with increasing output length (App.~Tab.~\ref{tab:slr_length_bins}), suggesting that extended reasoning traces often lead to derailment instead of deeper reasoning.

\section{Conclusion and Future Direction}
\label{sec:conclusion}

In this work, we introduced \slr for scalable and fully automated synthesis for logical reasoning benchmarks with verifiable rewards provided via logic programs. Our instantiation, \benchmark, offers a 20-level curriculum spanning 19k tasks with increasing logical complexity.

Our evaluations reveal that while current LLMs readily produce syntactically valid logic rules, robust logical reasoning remains elusive for conventional LLMs, especially as task complexity scales. Scaling parameters yield limited gains. Reasoning LLMs, aided by increased test-time compute, close part of this gap, albeit at significant computational costs.
Notably, curriculum learning on \benchmark substantially boosts in-domain and downstream reasoning. Specifically, our \slr-tuned Llama3-8B outperforms most conventional LLMs on \benchmark and surpasses several SOTA reasoning LLMs at a fraction of their inference costs.

Looking ahead, \slr enables diverse extensions, including reinforcement learning, richer logical domains, benchmarking neuro-symbolic and interactive reasoning systems, and advancing toward higher-order tasks such as causal inference. While \benchmark intentionally focuses on clean, single-rule ILP tasks with a perfect symbolic judge to isolate reasoning ability, future work can incorporate controlled forms of real-world noise, ambiguity, and multi-rule interactions to systematically study robustness alongside reasoning capabilities. Overall, \slr provides a flexible and extensible testbed for probing and improving LLM reasoning.

\section{Limitations}
\label{sec:limitations}

While \slr{} and \benchmark{} provide a scalable testbed for logical reasoning, there remain many opportunities for further enrichment. Although \benchmark{} currently applies \slr{} to the train domain with a single rule, the framework is readily extensible to multiple more complex, multi-rule reasoning scenarios and to entirely different domains. Our current focus on first-order, function-free Horn clauses enables systematic benchmark creation and evaluation; future instantiations could expand towards higher-order logic or probabilistic reasoning. While synthetic task generation comes with many benefits, such as ensuring novelty and precise control, it makes it difficult to incorporate real-world diversity and ambiguity. Our symbolic judge provides deterministic, discrete scoring and could potentially be enhanced to also recognize partial solutions, syntactically invalid rules, or natural language formulations. Overall, these points highlight the flexibility of our framework and outline promising directions for broadening its reach and impact.

\section*{Acknowledgments} We acknowledge support of the hessian.AI Innovation Lab (funded by the Federal Ministry of Research, Technology and Space, BMFTR, grant no. 16IS22091), the hessian.AISC Service Center (funded by the Federal Ministry of Education and Research, BMBF,  grant No 01IS22091), and the Center for European Research in Trusted AI (CERTAIN). Further, this work benefited from the ICT-48 Network of AI Research Excellence Center ``TAILOR'' (EU Horizon 2020, GA No 952215), the Hessian research priority program LOEWE within the project “WhiteBox”, the HMWK cluster projects ``Adaptive Mind'' and ``Third Wave of AI'', and from the NHR4CES. This work has also benefited from the BMWE project "Sovereign Open Source Foundational Models für European Intelligence (SOOFI)," 13IPC040G, and also from early stages of the Cluster of Excellence "Reasonable AI" funded by the German Research Foundation (DFG) under Germany’s Excellence Strategy— EXC-3057; funding will begin in 2026. This work was supported by the Priority Program (SPP) 2422 in the subproject “Optimization of active surface design of high-speed progressive tools using machine and deep learning algorithms“ funded by the German Research Foundation (DFG). Further, this work was funded by the European Union (Grant Agreement no. 101120763 - TANGO) as well as the AlephAlpha Collaboration lab 1141. This work was supported in part by OpenAI Research Credits.

\section*{Broader Impact}

\slr and \benchmark provide a scalable, reproducible foundation for evaluating and advancing logical reasoning in AI without relying on human annotation. By enabling robust measurement and targeted training, our framework supports progress in areas such as scientific discovery, program synthesis, and trustworthy AI. However, as LLMs acquire deeper logical competence, the risk of dual-use knowledge increases, enabling beneficial applications but also the potential for misuse, such as generating deceptive arguments or bypassing safety mechanisms. We urge responsible use and active consideration of ethical risks as these capabilities advance.

\bibliography{bibliography}

\clearpage

\appendix
\definecolor{codegreen}{rgb}{0,0.6,0}
\definecolor{codegray}{rgb}{0.5,0.5,0.5}
\definecolor{codepurple}{rgb}{0.58,0,0.82}
\definecolor{backcolour}{rgb}{0.95,0.95,0.92}

\newcommand{\reason}{\ding{51}} 
\newcommand{\noreason}{\ding{55}} 

\newcommand{\logosize}{0.9em}
\newcommand{\openai}{\raisebox{-0.15em}{\includegraphics[height=\logosize]{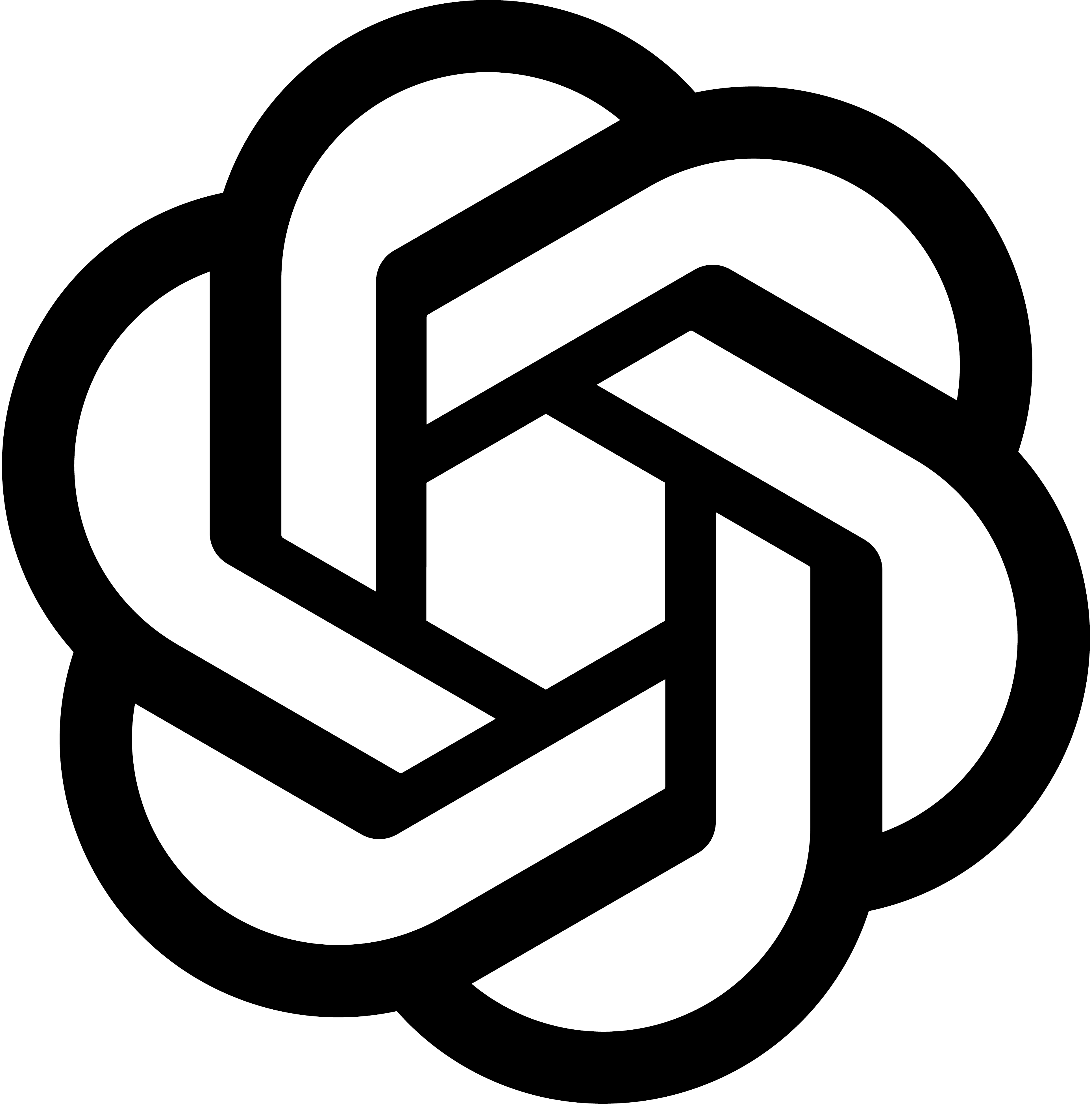}}}
\newcommand{\google}{\raisebox{-0.15em}{\includegraphics[height=\logosize]{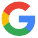}}}
\newcommand{\qwen}{\raisebox{-0.15em}{\includegraphics[height=\logosize]{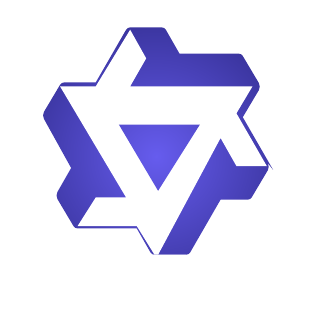}}}
\newcommand{\meta}{\raisebox{-0.15em}{\includegraphics[height=\logosize]{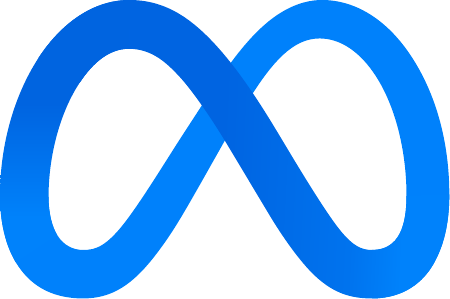}}}
\newcommand{\deepseek}{\raisebox{-0.15em}{\includegraphics[height=\logosize]{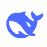}}}

\lstdefinestyle{mystyle}{
    backgroundcolor=\color{backcolour},
    commentstyle=\color{codegreen},
    keywordstyle=\color{magenta},
    numberstyle=\tiny\color{codegray},
    stringstyle=\color{codepurple},
    basicstyle=\footnotesize\ttfamily,
    breakatwhitespace=false,
    breaklines=true,
    captionpos=b,
    keepspaces=true,
    numbers=left,
    numbersep=5pt,
    showspaces=false,
    showstringspaces=false,
    showtabs=false,
    tabsize=2
}

\lstset{style=mystyle}

\begin{table*}[t!]
  \setlength{\tabcolsep}{2.5pt}
  \renewcommand{\arraystretch}{1}
  \centering
  \caption{
    \textbf{\benchmark Learning Curriculum}: The table details how both language and task configuration  are systematically increased throughout the curriculum levels. Notably, higher levels involve richer problems with more constants and predicates, larger problems, longer rules, and a transition from mirror to uniform and LLM-guided sampling. The final column reports the approximate combinatorial size of unique tasks available at each level.
  }
  \resizebox{\textwidth}{!}{
  \begin{tabular}{
    l
    >{\columncolor{mygrey!20}}c >{\columncolor{mygrey!20}}c >{\columncolor{mygrey!20}}c >{\columncolor{mygrey!20}}c >{\columncolor{mygrey!20}}c
    >{\columncolor{mygreen!15}}c >{\columncolor{mygreen!15}}c >{\columncolor{mygreen!15}}c >{\columncolor{mygreen!15}}c >{\columncolor{mygreen!15}}c
    >{\columncolor{myblue!15}}c >{\columncolor{myblue!15}}c >{\columncolor{myblue!15}}c >{\columncolor{myblue!15}}c >{\columncolor{myblue!15}}c
    >{\columncolor{mypurple!10}}c >{\columncolor{mypurple!10}}c >{\columncolor{mypurple!10}}c >{\columncolor{mypurple!10}}c >{\columncolor{mypurple!10}}c
  }
  \hline
    Stage &
      \multicolumn{5}{c}{\cellcolor{mygrey!50}Basic} &
      \multicolumn{5}{c}{\cellcolor{mygreen!50}Easy} &
      \multicolumn{5}{c}{\cellcolor{myblue!50}Medium} &
      \multicolumn{5}{c}{\cellcolor{mypurple!50}Hard} \\
    Levels & 1 & 2 & 3 & 4 & 5 & 6 & 7 & 8 & 9 & 10 & 11 & 12 & 13 & 14 & 15 & 16 & 17 & 18 & 19 & 20 \\
    \hline
    Language &&&&&&&&&&&&&&&&&&&&\\
    \quad \#Const.      & 1 & 1 & 1 & 2 & 2 & 2 & 2 & 2-3 & 2-3 & 2-3 & 2-4 & 2-4 & 4-6 & 4-6 & 4-6 & 5-6 & 5-6 & 5-6 & 5-6 & 5-6 \\
    \quad \#Pred.       & 5 & 5 & 5 & 5 & 5 & 5 & 6 & 6 & 6 & 7 & 7 & 9 & 9 & 9 & 9 & 10 & 10 & 12 & 12 & 12 \\
    Task Config. &&&&&&&&&&&&&&&&&&&& \\
    \quad $\kappa$      & 2 & 2 & 4 & 4 & 6 & 6 & 6 & 8 & 10 & 12 & 14 & 16 & 18 & 20 & 22 & 24 & 26 & 28 & 30 & 32 \\
    \quad $B_\pi$       & $\mathcal{M}$ & $\mathcal{M}$ & $\mathcal{M}$ & $\mathcal{M}$ & $\mathcal{M}$ & $\mathcal{U}$ & $\mathcal{U}$ & $\mathcal{U}$ & $\mathcal{U}$ & $\mathcal{U}$ & $\mathcal{U}$ & $\mathcal{U}$ & $\mathcal{U}$ & $\mathcal{U}$ & $\mathcal{U}$ & $\mathcal{U}$ & $\mathcal{U}$ & $\mathcal{U}$ & $\mathcal{U}$ & $\mathcal{U}$ \\
    \quad $R_{len}$     & 1 & 1-2 & 1-2 & 1-2 & 1-2 & 1-2 & 1-2 & 1-2 & 2-3 & 2-3 & 2-3 & 3-4 & 3-4 & 4-5 & 4-5 & 4-5 & 4-5 & 4-5 & 5 & 5 \\
    \quad $R_{sample}$  & $\mathcal{U}$ & $\mathcal{U}$ & $\mathcal{U}$ & $\mathcal{U}$ & $\mathcal{U}$ &
      $\mathcal{U}/L$ & $\mathcal{U}/L$ & $\mathcal{U}/L$ & $\mathcal{U}/L$ & $\mathcal{U}/L$ &
      $\mathcal{U}/L$ & $\mathcal{U}/L$ & $\mathcal{U}/L$ & $\mathcal{U}/L$ & $\mathcal{U}/L$ &
      $\mathcal{U}/L$ & $\mathcal{U}/L$ & $\mathcal{U}/L$ & $\mathcal{U}/L$ & $\mathcal{U}/L$ \\
     \textit{Comb. Size} & $10^3$& $10^3$& $10^5$& $10^{10}$& $10^{16}$& $10^{16}$& $10^{24}$& $10^{32}$& $10^{40}$& $10^{55}$& $10^{65}$& $10^{120}$& $10^{271}$& $10^{300}$& $10^{330}$& $10^{507}$& $10^{549}$& $10^{805}$& $10^{861}$& $10^{919}$ \\
    \hline
  \end{tabular}
  }
  \label{app:tab:task_spec}
\begin{minipage}{\textwidth}
\footnotesize
$\mathcal{M}$: mirror sampling \phantom{00}$\mathcal{U}$: uniform sampling\phantom{00}$L$: LLM-guided generation
\end{minipage}
\end{table*}

\section{Extended Notation}
\label{app:sec:fol}
We revisit essential definitions of first-order logic that we follow in this paper.
An FOL {\it Language} $\mathcal{L}$ is a tuple $(\mathcal{P}, \mathcal{A}, \mathcal{F}, \mathcal{V})$,
where $\mathcal{P}$ is a set of predicates, $\mathcal{A}$ is a set o constants, $\mathcal{F}$ is a set of function symbols (functors), and $\mathcal{V}$ is a set of variables.
A {\it term} is a constant, a variable, or a term that consists of a functor.
A {\it ground term} is a term with no variables.
We denote $n$-ary predicate ${\tt p}$ by ${\tt p}/n$.
An {\it atom} is a formula ${\tt p(t_1, \ldots, t_n) }$, where ${\tt p}$ is an $n$-ary predicate symbol and ${\tt t_1, \ldots, t_n}$ are terms.
A {\it ground atom} or simply a {\it fact} is an atom with no variables.
A {\it literal} is an atom or its negation.
A {\it positive literal} is just an atom. 
A {\it negative literal} is the negation of an atom.
A {\it clause} is a finite disjunction ($\lor$) of literals. 
A {\it definite clause} is a clause with exactly one positive literal.
If  $A, B_1, \ldots, B_n$ are atoms, then $ A \lor \lnot B_1 \lor \ldots \lor \lnot B_n$ is a definite clause.
We write definite clauses in the form of $A~\mbox{:-}~B_1,\ldots,B_n$.
Atom $A$ is called the {\it head}, and set of negative atoms $\{B_1, \ldots, B_n\}$ is called the {\it body}.
We call definite clauses by \emph{rules} for simplicity in this paper.
An atom is an atomic \emph{formula}. For formula $F$ and $G$, $\lnot F$, $F \land G$, and $F \lor G$ are also formulas.
\emph{Interpretation} of language $\mathcal{L}$ is a tuple $(\mathcal{D}, \mathcal{I}_\mathcal{A}, \mathcal{I}_\mathcal{F}, \mathcal{I}_\mathcal{P})$, 
where $\mathcal{D}$ is the  domain, $\mathcal{I}_\mathcal{A}$ is the assignments of an element in $\mathcal{D}$ for each constant ${\tt a} \in \mathcal{A}$,
$\mathcal{I}_\mathcal{F}$ is the assignments of a function from $\mathcal{D}^n$ to $\mathcal{D}$ for each $n$-ary function symbol ${\tt f} \in \mathcal{F}$, 
and $\mathcal{I}_\mathcal{P}$ is the assignments of a function from $\mathcal{D}^n$ to $\{ \top, \bot \}$ for each $n$-ary predicate ${\tt p} \in \mathcal{P}$.
For language $\mathcal{L}$ and formula $X$, an interpretation $\mathcal{I}$ is a \emph{model} if the truth value of $X$ w.r.t $\mathcal{I}$ is true.
Formula $X$ is a \emph{logical consequence} or \emph{logical entailment} of a set of formulas $\mathcal{H}$, denoted $\mathcal{H} \models X$, if, $\mathcal{I}$ is a model for $\mathcal{H}$ implies that $\mathcal{I}$ is a model for $X$ for every interpretation $\mathcal{I}$ of $\mathcal{L}$.

\section{\benchmark Details} \label{app:sec:task_spec}

\subsection{Task Specification}
Each task in \benchmark{} is precisely governed by a combination of language features and task configuration parameters, enabling fine-grained control over complexity and diversity. A task specification comprises two main components: (i) the logical language, which determines the set of predicates and argument types available, and (ii) the task configuration, which defines structural aspects of the task such as problem size, background knowledge sampling, rule length, sampling strategy, and the combinatorial space of realizable tasks. Tab.~\ref{app:tab:task_spec} details the curriculum’s level-wise specifications, showing how both language elements and the task config to create the individual levels.

\subsection{Language} \label{sec:app:grammar}

\paragraph{Predicates and Types.}
\slr defines a flexible, extensible vocabulary to support the systematic generation and evaluation of logical reasoning tasks. The primary predicate signatures and their argument types used in \benchmark{} are:
\texttt{eastbound}(\texttt{TRAIN}),
\texttt{westbound}(\texttt{TRAIN}),
\texttt{has\_car}(\texttt{TRAIN},\texttt{CAR}),
\texttt{car\_num}(\texttt{CAR},\texttt{NUM}),
\texttt{car\_color}(\texttt{CAR},\texttt{COLOR}),
\texttt{car\_len}(\texttt{CAR},\texttt{LEN}),
\texttt{has\_wall}(\texttt{CAR},\texttt{WALL}),
\texttt{has\_roof}(\texttt{CAR},\texttt{ROOF}),
\texttt{has\_payload}(\texttt{CAR},\texttt{LOADS}),
\texttt{load\_num}(\texttt{CAR},\texttt{NPAY}),
\texttt{has\_wheel}(\texttt{CAR},\texttt{WHEELS}),
\texttt{has\_window}(\texttt{CAR},\texttt{WINDOW}),
\texttt{car\_type}(\texttt{CAR},\texttt{CTYPE}),

\paragraph{Grounding Domains.}
Each argument type is grounded in a finite set of discrete constants:
\begin{align*}
  \mathit{NUM}     &::= [0\texttt{-}9]+\\
  \mathit{CAR}     &::= [0\texttt{-}9]+\\
  \mathit{COLOR}   &::= \texttt{red}\mid\texttt{blue}\mid\texttt{green}\mid \\
  &\texttt{yellow}\mid\texttt{white}\\
  \mathit{LEN}     &::= \texttt{short}\mid\texttt{long}\\
  \mathit{WALL}    &::= \texttt{full}\mid\texttt{railing}\\
  \mathit{ROOF}    &::= \texttt{roof\_foundation}\mid \\
  &\texttt{solid\_roof}\mid \texttt{braced\_roof}\mid \\
  &\texttt{peaked\_roof}\mid\texttt{none}\\
  \mathit{WHEELS}  &::= \texttt{2}\mid\texttt{3}\\
  \mathit{LOADS}   &::= \texttt{blue\_box}\mid\texttt{golden\_vase}\mid \\ &\texttt{barrel}\mid\texttt{diamond}\mid\texttt{metal\_pot}\mid \\
  &\texttt{oval\_vase}\mid\texttt{none}\\
  \mathit{NPAY}    &::= \texttt{0}\mid\texttt{1}\mid\texttt{2}\mid\texttt{3}\\
  \mathit{WINDOW}  &::= \texttt{full}\mid\texttt{half}\mid\texttt{none}\\
  \mathit{CTYPE}   &::= \texttt{passenger}\mid\texttt{freight}\mid\texttt{mixed}\\
  \mathit{NPAX}    &::= [0\texttt{-}9]
\end{align*}

\paragraph{Grammar Constraints.} Predicates are only instantiated with semantically compatible constant types. For example, $\texttt{car\_color}(\cdot,\cdot)$ only takes car objects and color constants as arguments; ill-typed facts are excluded during synthesis.

\paragraph{Task Synthesis Algorithm.} Algorithm \ref{alg:synth} details the synthesis pipeline, which first generates a governing rule before iteratively sampling background contexts and queries. These candidates undergo stratified rejection sampling to curate balanced positive and negative sets of samples that compose the final task prompt and validation program.

\subsection{Learning Curriculum in Detail}\label{app:sec:curriculum}
\benchmark presents a 20-level curriculum that progresses systematically in difficulty and is organized into four overarching complexity tiers: \textit{basic}, \textit{easy}, \textit{medium}, and \textit{hard}. For a detailed overview of the level-wise task specifications, see App. Tab. \ref{app:tab:task_spec}. Each level increases in difficulty through systematic modifications to both the task language and configuration, including: (i) increasing the overall size of the task ($\kappa$), (ii) expanding the vocabulary by adding new car constants per train, and (iii) broadening the set of predicates and grounding domains. Additionally, (iv) we adapt the policy used for synthesizing background knowledge, (v) alter the length $R_{len}$ of the ground truth rule, and (vi) adjust the rule sampling policy. Finally, (vii) the combinatorial space of tasks, quantified as the approximate number of distinct tasks that can be generated per level, also increases across the curriculum.
The compounding effect of larger problems, expanded vocabulary (constants and predicates), and more complex rules not only yields exponential growth in the combinatorial space but also effectively increases the complexity of the tasks. Consequently, while models identifying surface-level patterns might solve early levels, reasoning in later levels becomes more sophisticated, demanding more advanced problem-solving capabilities to progress.

\subsection{Verifiable Logic Rewards}\label{app:sec:logicrewards}

The \textsc{SymbolicJudge} computes verifiable logic rewards for a candidate hypothesis $H$ with respect to a background knowledge base $B$ and sets of positive ($E^+$) and negative ($E^-$) examples. These rewards are used both for evaluation and for training, and they follow standard practice in Inductive Logic Programming (ILP). Concretely, the judge deterministically tests whether a hypothesis satisfies classical ILP coverage criteria via Prolog execution, which provides the operational semantics for entailment and refutation. A key advantage of this setup is that it naturally supports \emph{non-unique solutions}: although tasks in \benchmark are generated from a single ground-truth rule, the learned hypothesis is not required to match it syntactically. Multiple distinct hypotheses may be semantically equivalent in that they entail all positives and reject all negatives under $B$, and any such hypothesis is considered correct. This allows evaluation to focus purely on semantic correctness rather than surface form. We compute three distinct metrics:

\textbf{Syntax Validity Score:} This binary score (0 or 1) indicates whether the candidate hypothesis $H$ is a syntactically and semantically valid Prolog rule. This serves as a prerequisite check for further evaluation; if $H$ is invalid, other scores are typically assigned 0.

\textbf{$\textsc{OverallScore}$:} This metric provides a binary indication of perfect task completion. It is 1 if and only if the hypothesis $H$, in conjunction with the background knowledge $B$, correctly entails all positive examples ($E^+$) and correctly refutes all negative examples ($E^-$). Otherwise, the score is 0.
\begin{multline*}
\textsc{OverallScore}_{B, E^+, E^-}(H) = \llbracket \\
\forall q\in E^+: (B \cup H) \models q \quad \land \\
\forall q\in E^-: (B \cup H) \not\models q \rrbracket \in \{0,1\}
\end{multline*}
Where $\llbracket \cdot \rrbracket$ is the Iverson bracket, evaluating to 1 if the condition inside is true, and 0 otherwise.

\textbf{$\textsc{PartialScore}$:} This metric reflects the fraction of examples (from both $E^+$ and $E^-$) that are correctly classified by the candidate hypothesis $H$ when combined with the background knowledge $B$. This provides a continuous signal of progress, even when the overall task is not perfectly completed.
\begin{multline*}
\textsc{PartialScore}_{B, E^+, E^-}(H)= \\ 
\frac{
      \sum_{q\in E^{+}}
      \llbracket (B\cup H)\models q \rrbracket
      + 
        \sum_{q\in E^{-}}
      \llbracket (B\cup H) \not\models q \rrbracket}{|E^{+}\cup E^{-}|}
\end{multline*}
Here, the numerator sums the count of correctly entailed positive examples and correctly not entailed negative examples. The denominator is the total number of examples, ensuring the score is normalized between 0 and 1.

These metrics provide rich feedback for both discrete evaluation (e.g., for filtering valid rules) and continuous optimization (e.g., for guiding reinforcement learning agents), allowing for robust assessment of learned logical hypotheses.

\subsection{LLM-Guided Rule Generation}\label{sec:llmprompt}
This section provides the prompt used for LLM-guided rule generation. The prompt was carefully designed to be both diverse and comprehensive, including a wide range of logical structures and Prolog features such as conjunction, disjunction, negation, recursion, aggregation, and pattern matching. By presenting the model with these varied and complex examples, we encourage the generation of challenging and realistic logic rules. This diversity is crucial for robust model generation of new rules, as it ensures that the LLM is exposed to representative samples of possible rule types encountered in real-world logic programming tasks.

\textbf{1. Conjunction with Existential Quantification: There exists a red short car}
There is at least one car that is both short and red.
\begin{lstlisting}[language=Prolog]
eastbound(Train) :-
    has_car(Train, Car),
    car_color(Car, red),
    car_len(Car, short).
\end{lstlisting}

\textbf{2. Disjunction: Some car is white or yellow.}
At least one car is either white or yellow.
\begin{lstlisting}[language=Prolog]
eastbound(Train) :-
    has_car(Train, Car),
    (car_color(Car, white) ; car_color(Car, yellow)).
\end{lstlisting}

\textbf{3. Negation: The train does not contain any red cars}
No car on the train is red.
\begin{lstlisting}[language=Prolog]
eastbound(Train) :-
    \+ (has_car(Train, Car), car_color(Car, red)).
\end{lstlisting}

\textbf{4. Inequality/Distinctness: Two cars must have different colors}
There are at least two cars on the train with different colors.
\begin{lstlisting}[language=Prolog]
eastbound(Train) :-
    has_car(Train, CarA),
    has_car(Train, CarB),
    CarA \= CarB,
    car_color(CarA, Color1),
    car_color(CarB, Color2),
    Color1 \= Color2.
\end{lstlisting}

\textbf{5. Aggregation/Counting: There are more green cars than yellow cars}
The train contains more green cars than yellow cars.
\begin{lstlisting}[language=Prolog]
eastbound(Train) :-
    findall(Car, (has_car(Train, Car), car_color(Car, green)), Greens),
    findall(Car, (has_car(Train, Car), car_color(Car, yellow)), Yellows),
    length(Greens, G),
    length(Yellows, Y),
    G > Y.
\end{lstlisting}

\textbf{6. Mutual Exclusion: Only one car is yellow; all others are not yellow}
There is exactly one yellow car; all others are not yellow.
\begin{lstlisting}[language=Prolog]
eastbound(Train) :-
    findall(Car, (has_car(Train, Car), car_color(Car, yellow)), [YellowCar]),
    forall(
        (has_car(Train, Car), Car \= YellowCar),
        (car_color(Car, NotYellow),
        NotYellow \= yellow)
    ).
\end{lstlisting}

\textbf{7. Uniqueness: No two cars have the same color}
All cars have unique colors.
\begin{lstlisting}[language=Prolog]
eastbound(Train) :-
    findall(Color, (has_car(Train, Car), car_color(Car, Color)), Colors),
    sort(Colors, UniqueColors),
    length(Colors, N),
    length(UniqueColors, N).
\end{lstlisting}

\textbf{8. No-Other/Uniqueness: Only two cars in the train}
Only two cars are present in the train.
\begin{lstlisting}[language=Prolog]
eastbound(Train) :-
    findall(Car, has_car(Train, Car), Cars),
    length(Cars, 2).
\end{lstlisting}

\textbf{9. Universal Quantification: Every full-wall car is long}
All cars with a full wall must be long.
\begin{lstlisting}[language=Prolog]
eastbound(Train) :-
    forall(
        (has_car(Train, Car), has_wall(Car, full)),
        car_len(Car, long)
    ).
\end{lstlisting}

\textbf{10. Conditional Implication: All long cars are either red or blue}
Every long car is either red or blue.
\begin{lstlisting}[language=Prolog]
eastbound(Train) :-
    forall(
        (has_car(Train, Car), car_len(Car, long)),
        (car_color(Car, Color), (Color = red ; Color = blue))
    ).
\end{lstlisting}

\textbf{11. Conditional Aggregation: All long cars are either red or blue}
Every long car is either red or blue.
\begin{lstlisting}[language=Prolog]
eastbound(Train) :-
    forall(
        (has_car(Train, Car), car_len(Car, long)),
        (car_color(Car, Color), (Color = red ; Color = blue))
    ).
\end{lstlisting}

\textbf{12. Pattern Matching: All full-wall cars are white}
Every full-wall car is white.
\begin{lstlisting}[language=Prolog]
eastbound(Train) :-
    forall(
      (has_car(Train, Car), has_wall(Car, full)),
      car_color(Car, white)
    ).
\end{lstlisting}

\textbf{13. Symmetry: Two cars are neighbors with same color}
CarA and CarB are neighbors on the train and have the same color.
\begin{lstlisting}[language=Prolog]
eastbound(Train) :-
    has_car(Train, CarA),
    has_car(Train, CarB),
    CarA \= CarB,
    car_num(CarA, N1),
    car_num(CarB, N2),
    (N2 =:= N1 + 1 ; N2 =:= N1 - 1),
    car_color(CarA, Color),
    car_color(CarB, Color).
\end{lstlisting}

\textbf{14. Combinatorial Group: Exactly two short yellow cars}
There are exactly two yellow cars, and both are short.
\begin{lstlisting}[language=Prolog]
eastbound(Train) :-
    findall(Car, (has_car(Train, Car), car_color(Car, yellow), car_len(Car, short)), L),
    length(L, 2).
\end{lstlisting}

\textbf{15. Recursion: At least one long car in the train}
The train has at least one long car.
\begin{lstlisting}[language=Prolog]
eastbound([Car|Cars]) :-
    car_len(Car, long)
    ;
    eastbound(Cars).
\end{lstlisting}

\textbf{16. Existence of a Structure (Sublist Pattern Matching)}
Exists three cars in sequence: Num, Num+1, Num+2, matching pattern.
\begin{lstlisting}[language=Prolog]
eastbound(Train) :-
    has_car(Train, Car1), car_num(Car1, N),
    car_len(Car1, short),
    N2 is N+1, N3 is N+2,
    has_car(Train, Car2), car_num(Car2, N2), car_len(Car2, long),
    has_car(Train, Car3), car_num(Car3, N3), car_len(Car3, short).
\end{lstlisting}

\textbf{17. Min/Max and Extremal Values}
A short car followed by a long car followed by a short car, anywhere in the train.
\begin{lstlisting}[language=Prolog]
eastbound(Train) :-
    findall(N, (has_car(Train, Car), car_num(Car, N)), Numbers),
    max_list(Numbers, Max),
    has_car(Train, LastCar),
    car_num(LastCar, Max),
    car_color(LastCar, white).
\end{lstlisting}

\textbf{18. Subset/Superset Constraints}
All full-wall cars are among the first three cars.
\begin{lstlisting}[language=Prolog]
eastbound(Train) :-
    forall(
      (has_car(Train, Car), has_wall(Car, full)),
      (car_num(Car, N), N =< 3)
    ).
\end{lstlisting}

\textbf{19. Projection/Aggregation Over Multiple Properties}
All pairs of cars have different (color, length) tuples.
\begin{lstlisting}[language=Prolog]
eastbound(Train) :-
    has_car(Train, CarA), has_car(Train, CarB), CarA \= CarB,
    car_color(CarA, ColA), car_len(CarA, LenA),
    car_color(CarB, ColB), car_len(CarB, LenB),
    (ColA \= ColB ; LenA \= LenB).
\end{lstlisting}

\textbf{20. All-Different on Multiple Attributes}
Enforce all car colors are different, AND all car numbers are different (car numbers are unique by assumption, but see structure).
\begin{lstlisting}[language=Prolog]
eastbound(Train) :-
    findall(Color, (has_car(Train, Car), car_color(Car, Color)), Colors),
    sort(Colors, UniqueColors),
    length(Colors, N), length(UniqueColors, N).
\end{lstlisting}

\subsection{Data Leakage Guarantees.}
We verify that no ground-truth rules ($R^\star$) from the training split reappear in the test split, next to the uniqueness of the background. While ensuring complete semantic independence is inherently challenging, surface-level similarity measures (e.g., predicate co-occurrence or template hashing) are insufficient to capture true logical overlap. Semantically distinct rules may share identical predicates and constants (e.g., “there exists a red car” vs. “all cars are red”).

\begin{table*}[t]
\centering
\resizebox{0.99\textwidth}{!}{
\begin{tabular}{l >{\tiny}l rrrl}
\toprule
\textbf{Model} & \textbf{Model Tag} & \textbf{Input} & \textbf{Input (Cached)} & \textbf{Output} & \textbf{API} \\
\midrule
\textbf{OpenAI Models} \\
gpt-5                  & gpt-5                          & 1.25   & 1.25   & 10.00  & OpenAI \\
gpt-5-mini             & gpt-5-mini                     & 0.25   & 0.25   & 2.00   & OpenAI \\
gpt-5-nano             & gpt-5-nano                     & 0.05   & 0.005  & 0.40   & OpenAI \\
gpt-4.5-preview        & gpt-4.5-preview-2025-02-27     & 75.00  & 37.50  & 150.00 & OpenAI \\
gpt-4o                 & gpt-4o-2024-08-06              & 2.50   & 1.25   & 10.00  & OpenAI \\
o1                     & o1-2024-12-17                  & 15.00  & 7.50   & 60.00  & OpenAI \\
o3                     & o3-2025-04-16                  & 10.00  & 2.50   & 40.00  & OpenAI \\
o4-mini                & o4-mini-2025-04-16             & 1.10   & 0.275  & 4.40   & OpenAI \\
o3-mini                & o3-mini-2025-01-31             & 1.10   & 0.55   & 4.40   & OpenAI \\
o1-mini                & o1-mini-2024-09-12             & 1.10   & 0.55   & 4.40   & OpenAI \\
\midrule
\textbf{Open-Source Models} \\
Qwen3-0.6B             & Qwen3-0.6B                     & 0.015  &        & 0.025  & OpenRouter \\
Qwen3-1.7B             & Qwen3-1.7B                     & 0.015  &        & 0.025  & OpenRouter \\
Qwen3-4B               & Qwen3-4B                       & 0.015  &        & 0.025  & OpenRouter \\
Qwen3-8B               & Qwen3-8B                       & 0.035  &        & 0.138  & OpenRouter \\
Qwen3-14B              & Qwen3-14B                      & 0.05   &        & 0.22   & OpenRouter \\
Qwen3-32B              & Qwen3-32B                      & 0.05   &        & 0.20   & OpenRouter \\
Qwen3-235B             & Qwen3-235B                     & 0.18   &        & 0.54   & OpenRouter \\
Qwen3-30B-A3B          & Qwen3-30B-A3B                  & 0.06   &        & 0.22   & OpenRouter \\
Qwen3-Coder-30B-A3B    & Qwen3-Coder-30B-A3B            & 0.06   &        & 0.25   & OpenRouter \\
Qwen3-Coder-408B-A35B  & Qwen3-Coder-408B-A35B          & 0.38   &        & 1.53   & OpenRouter \\
Qwen3-Next-80B-A3B & Qwen3-Next-80B-A3B-Instruct     & 0.10   &        & 0.80   & OpenRouter \\
& Qwen3-Next-80B-A3B-Thinking     & 0.15   &        & 1.20   & OpenRouter \\
\midrule
\textbf{Open-Source VLMs} \\
Qwen3-VL-2B            & Qwen3-VL-2B                    & 0.00   &        & 0.00   & OpenRouter \\
Qwen3-VL-4B            & Qwen3-VL-4B                    & 0.00   &        & 0.00   & OpenRouter \\
Qwen3-VL-8B            & Qwen3-VL-8B                    & 0.008  &        & 0.50   & OpenRouter \\
Qwen3-VL-30B           & Qwen3-VL-30B                   & 0.15   &        & 0.60   & OpenRouter \\
Qwen3-VL-32B           & Qwen3-VL-32B                   & 0.35   &        & 1.10   & OpenRouter \\
Qwen3-VL-235B          & Qwen3-VL-235B                  & 0.22   &        & 0.88   & OpenRouter \\
\bottomrule
\end{tabular}
}
\caption{Model Pricing (\$ per 1M tokens). API rates as of 01.05.2025, with Qwen3 and GPT5 model pricing added on 01.12.2025}
\label{tab:pricing}
\end{table*}
\subsection{\benchmark Task Synthesis}
App.~Fig.~\ref{app:fig:synthesis} illustrates a full synthesis run of \benchmark{} according to Alg.~\ref{alg:synth}. Starting from a specified language and configuration (rule length, problem size), the synthesizer first generates a latent ground-truth rule, then iteratively samples background facts until the desired number of positive and negative examples is met. The resulting task instance $\mathcal{I} = (B, E^+, E^-)$ forms a self-contained reasoning problem, which can be expressed either symbolically (Prolog format) or in natural language. This process demonstrates how \slr produces logically consistent, verifiable tasks with controllable difficulty and interpretable representations.

\section{Compute Costs and Model Pricing} \label{sec:app:apicosts}
\paragraph{Compute Costs.} Compute costs are reported as the total USD cost to run all prompts, based on publicly listed API prices as of 01.05.2025, with Qwen3 and GPT5 model pricing added on 01.12.2025. Pricing ignores server-side token caching, as actual cache hit counts are unavailable. Table~\ref{tab:pricing} summarizes per-model cost rates and API sources.

\begin{table*}[t]
\centering
\small
\resizebox{\textwidth}{!}{
\begin{tabular}{p{2cm} p{4.4cm} p{4.6cm} p{3.2cm}}
\toprule
\textbf{Dataset} & \textbf{Scope \& Task} & \textbf{Reasoning Specialty} & \textbf{Reference} \\
\midrule
MMLU
& Multi-domain analytical reasoning
& Factual knowledge, mathematics, statistics, computer science
& \cite{hendryckstest2021} \\
CLUTRR
& Relational and multi-hop inference
& Structured family graphs
& \cite{sinha2019clutrr} \\

LogiQA (2.0)
& Formal deductive reasoning
& Argument consistency
& \cite{liu2020logiqa,Hanmeng2023logicqa2} \\

GPQA
& High-level scientific reasoning
& Graduate-level difficulty
& \cite{rein2024gpqa} \\

ARC
& Symbolic abstraction
& Pattern-based problem solving
& \cite{Clark2018ThinkYH} \\

HellaSwag
& Grounded commonsense inference
& Contextual continuity
& \cite{zellers2019hellaswag} \\
\bottomrule
\end{tabular}
}
\caption{\textbf{Downstream reasoning datasets.} Overview of downstream evaluation datasets and what they evaluate for.}
\label{app:tab:datasets}
\end{table*}
\section{Training and Evaluation Details}\label{sec:training_details}
\subsection{Training Setup}
For our experiments on \benchmark, we fine-tune Llama-3.1-8B-Instruct using supervised fine-tuning (SFT) with LoRA adapters using LLaMA-Factory \cite{zheng2024llamafactory}. Training is performed stage-wise, presenting basic, easy, and medium problems sequentially, reflecting a curriculum of increasing logical complexity. Training is distributed across 8 GPUs using DeepSpeed with ZeRO Stage 3 optimization, taking 4 hours. Both optimizer states and model parameters are offloaded to CPU with pinned memory to maximize GPU memory efficiency. The AdamW optimizer is used in conjunction with a Warmup Cosine learning rate scheduler. Mixed-precision training is employed, with both bfloat16 and fp16 enabled in automatic mode. Communication overlap and contiguous gradients are activated to improve throughput, and model weights are saved in 16-bit precision at each checkpoint. Due to memory limitations, input sequences are truncated to a maximum length of 6k tokens using the Llama3 template, restricting training to \texttt{slr\_basic\_train}, \texttt{slr\_easy\_train}, and \texttt{slr\_medium\_train} splits. Optimization is performed using cross-entropy loss over the ground truth rule $R^\star$, with a per-device batch size of 5 and gradient accumulation over 2 steps, resulting in an effective batch size of 80 samples per step across 8 GPUs. The learning rate is set to $2 \times 10^{-4}$, scheduled with a cosine scheduler and a warmup ratio of 0.03. All relevant hyperparameters and training scripts are included in the codebase for full reproducibility.
\subsection{Evaluation Setup}
To evaluate the generality of reasoning skills induced by \slr, we benchmark models on a diverse set of widely used reasoning datasets. These benchmarks span multiple reasoning modalities, including multi-domain analytical reasoning, relational and multi-hop inference, formal deductive reasoning, symbolic abstraction, scientific reasoning, and grounded commonsense inference. Together, they provide broad coverage of the reasoning capabilities relevant for modern large language models. An overview of all downstream datasets, including their scope, reasoning focus, and references, is provided in App.~Tab.~\ref{app:tab:datasets}.
MMLU and its subsets assess analytical reasoning across factual knowledge, mathematics, statistics, and computer science~\cite{hendryckstest2021}. CLUTRR evaluates relational and multi-hop inference over structured family graphs~\cite{sinha2019clutrr}. LogiQA and LogiQA2 focus on formal deductive reasoning and logical consistency in natural language arguments~\cite{liu2020logiqa,Hanmeng2023logicqa2}. GPQA and its variants (Extended and Diamond) measure graduate-level scientific reasoning~\cite{rein2024gpqa}. ARC-Easy and ARC-Challenge test symbolic abstraction and pattern-based reasoning~\cite{Clark2018ThinkYH}, while HellaSwag evaluates grounded commonsense inference and contextual continuity~\cite{zellers2019hellaswag}.
For downstream evaluation, we use the Language Model Evaluation Harness~\cite{evalharness} with the default protocol of each benchmark, enabling few-shot as multiturn prompting to support multi-turn contexts where applicable, including the official pass rate (typically pass@1) and whether evaluation is performed in zero-shot or few-shot mode. CLUTRR, which is not supported by evalharness, follows a similar setup with accuracy averaged across all test sets. All evaluations are conducted on 8 GPUs using vLLM~\cite{kwon2023efficient} for efficient batch inference. Reported scores reflect accuracy for each model and benchmark, and all results are based on the official evaluation splits and standardized prompt formatting consistent with the \slr curriculum.

\section{Ablation Studies}
\subsection{Ablation across synthesis parameters}\label{app:sec:synthesis_parameters}
\begin{table}[t]
\centering
\resizebox{\linewidth}{!}{
\begin{tabular}{lcr}
\toprule
\textbf{Ablation Factor} & \textbf{Configuration} & \textbf{Accuracy (\%)} \\
\midrule
\multirow{2}{*}{\textbf{Rule Sampling}} 
  & Uniform     & \textbf{11.0}\phantom{ (↓9.5)} \\
  & LLM-guided  & 2.5  {\color{red}(↓8.6)} \\
\midrule
\multirow{2}{*}{\textbf{Problem Size ($\boldsymbol{\kappa}$)}} 
  & $\kappa=4$  & \textbf{35.0}\phantom{ (↓9.5)}  \\
  & $\kappa=6$  & 30.0  {\color{red}(↓5.0)} \\
\midrule
\multirow{2}{*}{\textbf{ ($\boldsymbol{R_{\text{len}}}$)}} 
  & 1--2 literals & \textbf{7.2}\phantom{ (↓9.5)}\\
  & 2--3 literals & 2.6  {\color{red}(↓4.6)} \\
\midrule
 \textbf{Background} & Mirror   & \textbf{30.0}\phantom{ (↓9.5)} \\
 \textbf{Sampling} & Uniform  & 20.5 {\color{red}(↓9.5)} \\
\bottomrule
\end{tabular}
}
\caption{\textbf{Ablations Across Synthesis Parameters.}
Accuracy (\%) averaged across the LLaMA base models for different synthesis parameters. 
Each row corresponds to one ablation factor, comparing paired configurations. 
Arrows in red indicate decreases relative to the first configuration in each pair.}
\label{tab:slr_ablation_matrix}
\end{table}
To assess how synthesis configurations affect task difficulty, we conducted controlled ablations across the rule sampling policy, problem size, rule length, and background sampling policy, averaging results across all open-source LLaMA base models. As summarized in App. Tab.~\ref{tab:slr_ablation_matrix}, these experiments confirm that \slr yields interpretable and controllable gradients in logical complexity. For each factor, we selected representative levels of the dataset where the targeted parameter changes.

\textbf{(1) Rule sampling policy.} To measure the influence of rule generation, we compare subsets within the easy-difficulty tier (levels 6–10) containing either uniformly sampled versus LLM-guided rules. As shown in Table~\ref{tab:slr_ablation_matrix}, accuracy drops sharply from 11.0 \% $\rightarrow$ 2.5 \%, demonstrating that LLM-guided synthesis introduces more structured and semantically complex rules that substantially increase reasoning difficulty.

\textbf{(2) Problem size ($\boldsymbol{\kappa}$).} To isolate the effect of larger reasoning tasks, we compare level 4 to level 5, where $\kappa$ increases from 4 to 6. This change leads to a moderate decrease in accuracy (35.0 \% $\rightarrow$ 30.0 \%), showing that models begin to struggle as the relational scope and combinatorial complexity expand.

\textbf{(3) Rule length ($\boldsymbol{R_{\text{len}}}$).} To examine the effect of rule depth, we compare level 8 (rules with one–two literals) and level 9 (two–three literals). Accuracy drops markedly from 7.2 \% $\rightarrow$ 2.6 \%, confirming that even modestly longer reasoning chains substantially raise logical complexity and error rates.

\textbf{(4) Background sampling policy.} Finally, we compare level 5, which employs mirror sampling, to level 6, which uses uniform sampling over the background facts. Accuracy decreases from 30.0 \% $\rightarrow$ 20.5 \%, indicating that the mirror trains are significantly easier to solve than the uniformly sampled trains.

Together, these ablations highlight that the synthesis parameters in \slr modulate task difficulty, confirming that \benchmark provides a precise and controllable testbed for analyzing logical reasoning performance across model families.
\begin{table}[t]
\centering
\resizebox{\linewidth}{!}{
\begin{tabular}{lccccc}
\toprule
\textbf{Curriculum} & \textbf{LRL} & \textbf{Basic} & \textbf{Easy} & \textbf{Medium} & \textbf{Hard} \\
& \scriptsize{($\uparrow$0-20)} & \scriptsize{($\uparrow$\%)} & \scriptsize{($\uparrow$\%)} & \scriptsize{($\uparrow$\%)} & \scriptsize{($\uparrow$\%)} \\
\midrule
Ordered          & {6.01} & {88.7} & 28.4 & 3.3 & 0.2 \\
Random         & 5.99 & 88.1 & {28.5} & 3.6 & 0.05 \\
Reverse  & 5.93 & 86.8 & 27.7 & {4.5} & 0.05 \\
\bottomrule
\end{tabular}
}
\caption{\textbf{Curriculum Learning Order.} Comparison of curriculum learning orders (ordered, random, and reverse) under matched training budgets (levels~5–7) using llama3.1-8b-it. 
The ordered curriculum achieves the highest overall LRL.}
\label{tab:curriculum_ablation}
\end{table}
\subsection{Curriculum Learning Order}
To evaluate the effect of curriculum ordering on reasoning acquisition, we conducted a controlled ablation comparing standard curriculum learning, random order, and reverse curriculum training. All settings were matched for dataset, sample size, and total token budget, using the same difficulty levels (5–7). As shown in Tab.~\ref{tab:curriculum_ablation}, the curriculum-trained model achieves the highest overall Logical Reasoning Level (LRL) and more balanced performance across tiers, while random and reverse orders yield slightly lower LRL and less stable behavior across difficulty levels. These results indicate that a structured progression of reasoning difficulty facilitates smoother learning and generalization in \slr.
\begin{table}[t]
\centering
\resizebox{\linewidth}{!}{
\begin{tabular}{lrrrrrr}
\toprule
\textbf{Model Variant} & \textbf{LRL} & \textbf{Basic} & \textbf{Easy} & \textbf{Medium} & \textbf{Hard} & \textbf{Cost (\$)} \\
\midrule
\textit{o4-mini-low}  & 10.3 & 91 & 81 & 25 &  9 & 7.26 \\
\textit{o4-mini}      & 12.3 & 93 & 88 & 52 & 13 & 21.43 \\
\textit{o4-mini-high} & 12.8 & 98 & 96 & 40 & 21 & 24.24 \\
\bottomrule
\end{tabular}
}
\caption{\textbf{Scaling Test-time Compute within the \textit{o4-mini} Family.} 
Increasing compute improves reasoning ability but raises inference costs and does not yield uniform accuracy gains across all tiers.}
\label{app:tab:o4mini}
\end{table}
\subsection{Scaling Test-time compute with \textit{o4-mini}}\label{app:sec:scaling_compute}
Table~\ref{app:tab:o4mini} analyzes the impact of increasing test-time compute within the \textit{o4-mini} family.  
The \textit{high} variant achieves the highest overall LRL, outperforming both \textit{o4-mini} and \textit{o4-mini-low}, but incurs a cost increase of over 13\%. Notably, it performs worse on the \textit{medium} tier (40\% vs.~52\%), indicating diminishing returns from additional compute.  This supports the main text finding that higher inference cost does not guarantee proportional reasoning gains.

\subsection{Compute Increases with Task Complexity}
\begin{figure}[t!]
    \centering
    \includegraphics[width=\linewidth]{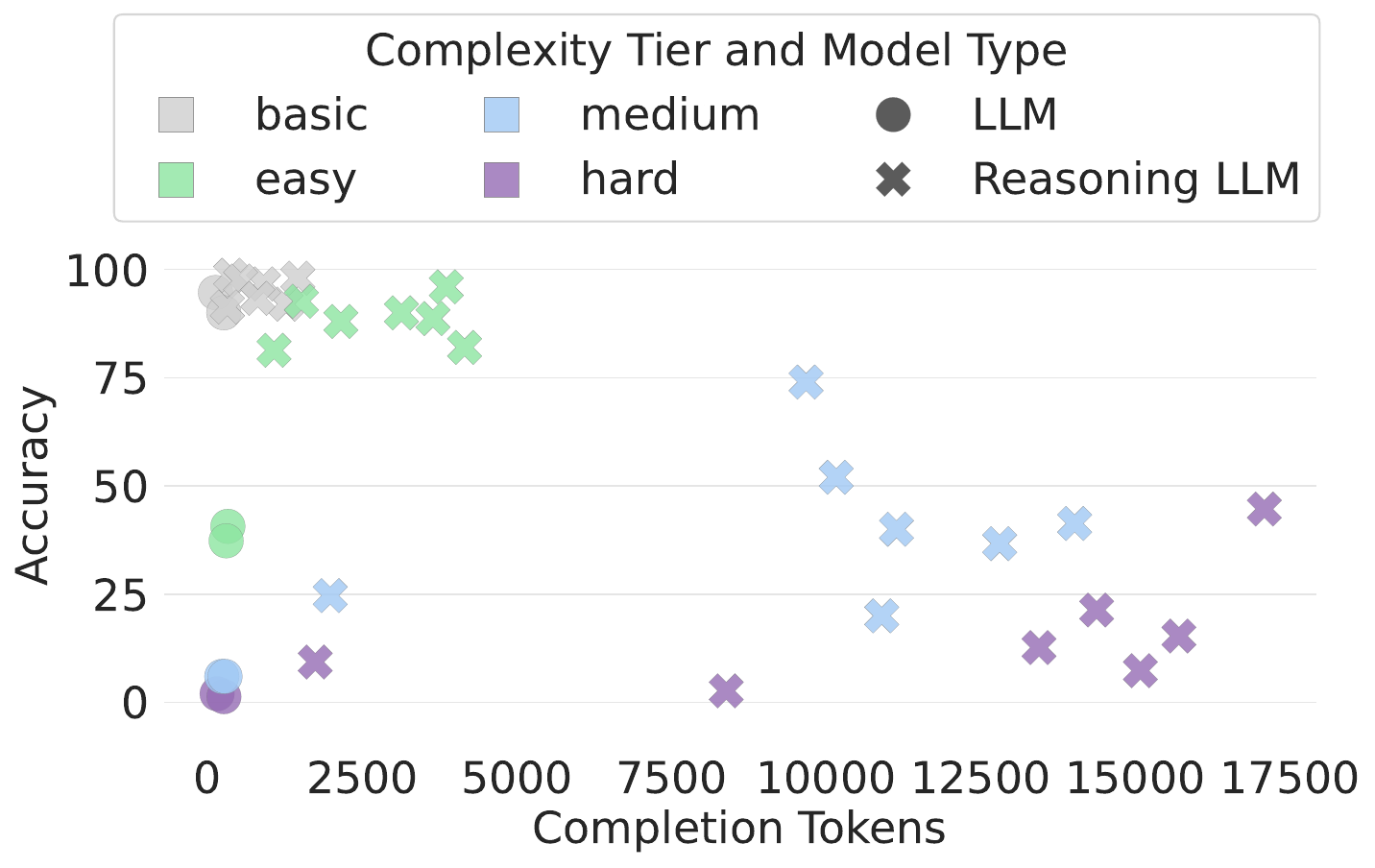}
    \caption{\textbf{Compute–Performance Trade-Off}. Reasoning LLMs achieve higher accuracy than base LLMs but require more compute. While more complex tasks typically demand more completion tokens, increased compute does not always translate to higher accuracy.
    }
    \label{fig:compute}
\end{figure}
As reasoning tasks grow in complexity, models generally require more completion tokens to solve them, leading to higher inference costs (see Fig.~\ref{fig:compute}). This trend serves as a sanity check confirming that \benchmark’s task difficulty correlates with increased reasoning effort rather than being arbitrary. However, since larger models inherently consume more compute per token, these results should not be interpreted as direct efficiency comparisons across models.

\begin{table*}[t]
    \centering
    \resizebox{\linewidth}{!}{
    \begin{tabular}{llllllllll@{}}
         & LRL \scriptsize{($\uparrow$0-20)}& Syntax\scriptsize{($\uparrow$\%)} & Basic\scriptsize{($\uparrow$\%)} & Easy\scriptsize{($\uparrow$\%)} & Medium\scriptsize{($\uparrow$\%)} & Hard\scriptsize{($\uparrow$\%)} \\
        \midrule
        Qwen3-8B & 4.6 & \phantom{0}93 & 79 & 14 & \phantom{0}0 & 0  \\
        \rowcolor{gray!15}\textbf{Qwen3-8B-SLR}
        & \textbf{7.9 }\textcolor{darkgreen}{\small{(+3.3)}}
        & \textbf{100 }\textcolor{darkgreen}{\small{(+7)}}
        & \textbf{96 }\textcolor{darkgreen}{\small{(+17)}}
        & \textbf{43 }\textcolor{darkgreen}{\small{(+29)}}
        & \textbf{14 }\textcolor{darkgreen}{\small{(+14)}}
        & \textbf{6 }\textcolor{darkgreen}{\small{(+6)}} \vspace{0.5em} \\  
        \multicolumn{7}{c}{\textit{Downstream Reasoning Performance}}\\
        & SLR-Homes {\scriptsize($\uparrow$\%)}& MMLU {\scriptsize($\uparrow$\%)}& MMLU-Stats {\scriptsize($\uparrow$\%)}& CLUTRR {\scriptsize($\uparrow$\%)}& LogiQA {\scriptsize($\uparrow$\%)}& LogiQA2 {\scriptsize($\uparrow$\%)}\\
        \midrule
       Qwen3-8B & 38.0 & 49.8 & 18.1  & 19.6 & 25.0 & 27.7 \\
       \rowcolor{gray!15}\textbf{Qwen3-8B-SLR} 
      & \textbf{40.7} \textcolor{darkgreen}{\small{(+2.7)}} 
       & \textbf{55.9} \textcolor{darkgreen}{\small{(+6.1)}} 
       & \textbf{18.1} \textcolor{gray}{\small{(+0.0)}} 
       & \textbf{20.6} \textcolor{darkgreen}{\small{(+1.0)}} 
       & \textbf{31.5} \textcolor{darkgreen}{\small{(+6.5)}} 
       & \textbf{31.0} \textcolor{darkgreen}{\small{(+3.3)}} \vspace{0.5em} \\
        & GPQA {\scriptsize($\uparrow$\%)}& GPQA-Ext. {\scriptsize($\uparrow$\%)}& GPQA-Dia. {\scriptsize($\uparrow$\%)}& ARC-Easy {\scriptsize($\uparrow$\%)}& ARC {\scriptsize($\uparrow$\%)}& HellaSwag {\scriptsize($\uparrow$\%)}\\
        \midrule
        Qwen3-8B & 23.4 & 26.4 & 21.2 & 69.2 & 43.3 & 48.8 \\
        \rowcolor{gray!15}\textbf{Qwen3-8B-SLR} 
        & \textbf{26.8} \textcolor{darkgreen}{\small{(+3.4)}} 
        & \textbf{27.1} \textcolor{darkgreen}{\small{(+0.7)}} 
        & \textbf{27.8} \textcolor{darkgreen}{\small{(+6.5)}} 
        & \textbf{77.3} \textcolor{darkgreen}{\small{(+8.1)}} 
        & \textbf{49.8} \textcolor{darkgreen}{\small{(+6.5)}} 
        & \textbf{51.6} \textcolor{darkgreen}{\small{(+2.8)}} \\
    \end{tabular}
        }
    \caption{
        \textbf{Qwen3-8B-SLR.}
        Benchmark scores {\scriptsize($\uparrow$)} for base and SLR-tuned models on \benchmark{} and downstream benchmarks; LRL measuring cumulative curriculum progress. The tuned model surpasses the baseline across all curriculum stages, while generalizing to other downstream reasoning tasks.
     }
     \label{app:tab:downstream_ext}
\end{table*}
\section{Additional Experiments}
\subsection{Extended Leaderboard}
We extend our evaluation to the Qwen3 and GPT-5 model families and consolidate all new model results into a single unified extended leaderboard (Tab.~\ref{tab:extended-leaderboard}). The leaderboard compares reasoning-oriented models (orange) with non-reasoning models (blue) across model families and scales, providing a comprehensive overview of current reasoning performance on SLR-Bench. The more recent models show improved LRL. Despite these advances, GPT-5 solves only 46\% of the tasks in the Hard tier, highlighting that SLR-Bench remains challenging even for the strongest contemporary models, particularly at higher levels of inductive complexity.
\subsection{Qwen3-SLR}
We fine-tune a LoRA adapter on Qwen3-8B using a similar setup as the llama model, but with SLR training data shuffled. Subsequently, we evaluate the resulting model on SLR-Bench and on downstream reasoning benchmarks. As shown in App.~Tab.~\ref{app:tab:downstream_ext}, Qwen-SLR consistently improves performance on SLR tasks and also yields gains across downstream reasoning benchmarks, indicating that the learned reasoning capabilities transfer beyond the training distribution.

\subsection{DeepSeek-SLR}
\begin{table*}[t]
    \centering
    \label{app:tab:qwen}
        \centering
        \resizebox{.9\linewidth}{!}{
        \begin{tabular}{llllllllll}
            \toprule
             & LRL \scriptsize{($\uparrow$0-20)}& Syntax\scriptsize{($\uparrow$\%)} & Basic\scriptsize{($\uparrow$\%)} & Easy\scriptsize{($\uparrow$\%)} & Medium\scriptsize{($\uparrow$\%)} & Hard\scriptsize{($\uparrow$\%)} \\
             \midrule         
            DeepSeek-R1-1.5B
            & \phantom{0.}0 & \phantom{0}54 & \phantom{0}0 & \phantom{0}0 & \phantom{0}0 & 0  \\
            \rowcolor{gray!15}\textbf{DeepSeek-1.5B-SLR}
            & \textbf{5.6 }\textcolor{darkgreen}{\small{(+5.6)}} 
            & \textbf{\phantom{0}97}\textcolor{darkgreen}{\small{(+43)}} 
            & \textbf{84 }\textcolor{darkgreen}{\small{(+84)}} 
            & \textbf{27 }\textcolor{darkgreen}{\small{(+27)}} 
            & \textbf{\phantom{0}2 }\textcolor{darkgreen}{\small{(+2)}}  
            & \textbf{0 }\textcolor{gray}{\small{(+0)}}   \\
            \bottomrule
    \end{tabular}
    }
    \caption{
        \textbf{DeepSeek-SLR.} GRPO training of DeepSeek-R1-1.5B on \benchmark. Benchmark scores {\scriptsize($\uparrow$\%)} for base and SLR-tuned models (DeepSeek-R1-1.5B) on \benchmark{}; LRL measures cumulative curriculum progress. The SLR-tuned model improves syntax and reasoning capabilities on basic, easy, and medium problems.
    }
\end{table*}
To examine whether the \slr curriculum remains effective in RL setups, we applied it using Group Reinforcement Preference Optimization (GRPO) \cite{Shao2024grpo} on the small-scale reasoning model \textit{deepseek-ai/DeepSeek-R1-Distill-Qwen-1.5B}. All training was performed under the same setup as SFT. The resulting model exhibits a clear improvement in reasoning ability, with the Logical Reasoning Level (LRL) increasing from 0.0 to 5.6. Performance gains are observed across reasoning tiers, especially on basic and easy reasoning tasks, demonstrating that a structured progression of task difficulty benefits even small models trained via reinforcement optimization. However, training was halted at level~7 due to reward collapse in later stages, suggesting that smaller architectures experience instability once the reasoning task complexity surpasses their capacity. Overall, these results indicate that the \slr curriculum not only supports SFT but also effectively enhances reasoning in RL setups (GRPO), reinforcing its flexibility as a general training paradigm.

\subsection{Extended Pass@k and CoT Results}\label{app:sec:cot}
\begin{table*}[t]
\centering
\resizebox{0.9\textwidth}{!}{
\begin{tabular}{lccccc}
\toprule
\textbf{Model} & LRL@1 \scriptsize{($\uparrow$0-20)} & LRL@4 \scriptsize{($\uparrow$0-20)} & LRL@8 \scriptsize{($\uparrow$0-20)} & Tokens \scriptsize{$(M\downarrow)$} & Cost \scriptsize{$(\$\downarrow$)} \\
\midrule
\rowcolor{gray!15} llama-3.1-8b-SLR & \textbf{8.2} & \textbf{10.5} & \textbf{11.3} & \textbf{0.05} & 0.14 \\
Llama-3.3-70B-IT-CoT           & 5.9 & 7.0 & 7.5 & 0.93 & 0.93 \\
Llama-3.3-70B-IT               & 5.6 & 6.8 & 7.3 & 0.49 & 0.82 \\
Llama-3.1-8B                   & 3.8 & 5.5 & 6.0 & 0.20 & 0.14 \\
Llama-3.1-8B-IT-CoT            & 3.4 & 5.3 & 6.0 & 1.37 & 0.18 \\
Llama-3.2-3B-IT-CoT            & 1.8 & 3.8 & 4.5 & 1.02 & 0.13 \\
Llama-3.2-3B-IT                & 1.0 & 2.6 & 3.6 & 0.10 & \textbf{0.11} \\
Llama-3.2-1B-IT-CoT            & 0.0 & 0.0 & 0.0 & 1.17 & 0.17 \\
Llama-3.2-1B-IT                & 0.0 & 0.0 & 0.0 & 0.47 & 0.12 \\
\bottomrule
\end{tabular}
}
\caption{\textbf{\benchmark Pass@k and CoT Results.} 
We report Logical Reasoning Level (LRL) scores at Pass@1, Pass@4, and Pass@8 (\%) for open-source models, together with total completion tokens (M) and inference cost (\$). 
Higher LRL values indicate stronger logical reasoning on \benchmark. 
Chain-of-Thought (CoT) variants illustrate the effect of explicit reasoning traces, while increasing $k$ reflects test-time sampling improvements.}
\label{tab:passk_cot_results}
\end{table*}
To complement the main results, where we report average Pass@1 (LRL) performance, we provide extended Pass@k scores (\textit{averaged for $k < 8$}) and Chain-of-Thought (CoT) results for open-source models in App.~Tab.~\ref{tab:passk_cot_results}. The table summarizes model performance at different test-time sampling levels ($k=1,4,8$), together with compute requirements in tokens and cost for a single pass. Across all models, accuracy increases with larger $k$, confirming that multiple sampled completions raise the probability of generating a correct reasoning trace. Beyond improving performance, Pass@k also reveals reasoning variance. Smaller models (e.g., 3B–8B) show larger Pass@1–Pass@8 gaps, doubling in accuracy as we move from pass@1 to pass@8, indicating less stable reasoning, while larger models display narrower gaps and more consistent inference. CoT variants generally show modest gains over their non-CoT counterparts of comparable scale, indicating that explicit reasoning traces can improve reasoning, but also come with higher computational costs. Overall, these findings show that both test-time sampling and CoT prompting not only enhance reasoning outcomes but also serve as useful probes into a model’s internal reasoning diversity and reliability.

\subsection{CLUTRR Evaluation}
\begin{table}[t]
\centering
\caption{\textbf{CLUTRR Evaluation.} Accuracy (\%) of base, SLR-tuned, and reasoning models on CLUTRR test sets (greedy decoding). SLR-tuned outperforms both base and reasoning models on CLUTRR.}
\resizebox{\linewidth}{!}{
\begin{tabular}{lccc}
\toprule
 & \multicolumn{3}{c}{{Llama-3.1-8B-it}} \\
\cmidrule(lr){2-4}
{CLUTRR Variant} & {Base} & {w/ SLR} & {w/ DeepSeek-R1} \\
\midrule
gen\_train23\_test2to10          & 10.2 & 19.1 & \textbf{24.0} \\
gen\_train234\_test2to10         & 9.5  & 16.4 & \textbf{25.0} \\
rob\_train\_clean\_23\_test\_all & 29.1 & \textbf{35.6} & 26.0 \\
rob\_train\_sup\_23\_test\_all   & 43.2 & \textbf{45.2} & 23.0 \\
rob\_train\_irr\_23\_test\_all   & 31.3 & \textbf{34.5} & 20.0 \\
rob\_train\_disc\_23\_test\_all  & 29.7 & \textbf{41.6} & 18.0 \\
\midrule
\textbf{Average}                 & 25.5 & \textbf{32.1} & 23.0 \\
\bottomrule
\end{tabular}
}
\label{app:tab:clutrr_eval}
\end{table}
We evaluated SLR-tuned and baseline models on the CLUTRR benchmark \cite{sinha2019clutrr}, an inductive reasoning dataset requiring  inference over kinship relations. All models were evaluated with greedy decoding, focusing on accuracy since CLUTRR uses a closed classification space where syntax validity does not apply. Across all variants, SLR-tuning consistently improves over the base model, with an average gain of +6.6pp, and remains competitive with or superior to DeepSeek-R1-8B on most configurations (App.~Tab.~\ref{app:tab:clutrr_eval}).
Notably, Llama-3.1-8B-SLR achieves higher accuracy than DeepSeek-R1-8B on four of six variants while using dramatically fewer resources—only 11k completion tokens vs. 8.1M for DeepSeek (730× less compute).
These results provide strong evidence that SLR enhances inductive reasoning beyond SLR-Bench, achieving higher accuracy and far greater computational efficiency on an established external inductive reasoning benchmark.

\section{Error Mode Analysis.} \label{app:sec:error_modes}

To better understand model failures on \benchmark, we conducted a comprehensive error analysis of generated rules. As discussed in Section \ref{sec:metabench}, syntax is rarely the limiting factor; most models achieve near-perfect syntactic validity. Consequently, the primary bottleneck lies in semantic rule induction.

\subsection{Taxonomy of Error Modes}

By inspecting a sample of failed generations across all models, we identified five prominent failure patterns:

\begin{enumerate}
    \item \textbf{Over-generalization:} Rules that are too broad, often consisting of only one or two predicates (e.g., $eastbound(T) :- has\_car(T,C).$).
    \item \textbf{Under-generalization:} Rules that are overly specific, often containing excessive literals (more than eight) that fail to capture the concise ground truth (e.g., $eastbound(T) :- has\_car(T,C), car\_len(C,short), \dots$).
    \item \textbf{Failure of Abstraction:} The model lacks semantic understanding of the induction task. Instead of lifting specific examples into variables, the model overfits to the prompt context, replacing general logic with grounded atoms (e.g., hardcoding $eastbound(T) :- car1$). This suggests the model is performing text completion on the prompt rather than logical reasoning.
    \item \textbf{Degenerate Reasoning:} The model falls into repetition loops, meaningless enumerations  (e.g., repeating “iiii”), or fails to produce structured logic entirely.
\end{enumerate}

\subsection{Quantitative analysis}

Quantitative analysis across all models reveals that 7\% of generations were short rules ($\leq$2 literals, 38\% success), 20\% overly long rules ($\geq$8 literals, 14\% success), 1\% had grounded heads (0\% success), and 8\% showed degenerate reasoning patterns (33\% success). Degenerations were identified via simple heuristics (e.g., $\geq10$ repeated characters, $\geq$10 enumerated lines, or $\geq30\%$ dominance of a single character). While approximate, these checks capture systematic breakdowns. Notably, even degenerate outputs sometimes partially recover, with 25\% still solving the task.

\subsection{Model-Specific Behaviors}
While most failure modes emerge across all models, some failure modes are predominantly assignable to some model classes.

\paragraph{Degenerate Reasoning in Distilled Reasoning Models.}
In contrast, distilled reasoning models (e.g., DeepSeek-R1-Llama-70B) struggle with \textit{Degenerate Reasoning} (Error Type 4). Unlike smaller models that simply fail to abstract, these larger distilled models attempt complex reasoning but frequently derail.
\begin{table}[t]
\centering
\caption{\textbf{Success Rate by Output Length Decile.}
Binning generations by relative output length (short $\rightarrow$ long) on DeepSeek-R1-Llama-80B. Longer reasoning traces correlate strongly with lower accuracy, indicating that verbosity often signals derailment rather than deeper reasoning.}
\resizebox{\linewidth}{!}{
\begin{tabular}{lcccccccccc}
\toprule
\textbf{Decile (short $\rightarrow$ long)} & 0 & 1 & 2 & 3 & 4 & 5 & 6 & 7 & 8 & 9 \\
\midrule
\textbf{Accuracy} (\%\scriptsize{$\uparrow$}) & 96 & 87 & 87 & 80 & 42 & 7 & 2 & 0 & 0 & 0 \\
\bottomrule
\end{tabular}
}
\label{tab:slr_length_bins}
\end{table}
As detailed in Table~\ref{tab:slr_length_bins}, we observed a strong inverse correlation between reasoning length and success. Success rates drop from 96\% in the shortest decile to 0\% in the longest, indicating verbosity reflects derailment rather than deeper reasoning.

\paragraph{Failure of Abstraction in Small Models.}
Conversely, smaller distilled models (e.g., Llama-3.2-1B) struggle with fundamental \textit{Failure of Abstraction} (Error Type 3). These models lack the capacity to maintain the distinction between the example data (constants) and the logical rule (variables). They rarely attempt shortcuts or fall into complex reasoning loops; instead, they simply parrot the input context or hallucinate grounded atoms, effectively treating the logic induction task as a text completion exercise.

\section{Code and Licenses}

This work introduces and publicly releases several scientific artifacts, including the \slr framework for scalable logical reasoning with large language models, the \benchmark dataset comprising 19{,}000 tasks across 20 curriculum levels, and associated training, evaluation, and logic validation scripts. All code and data with the logic reward interface will be made publicly available after publication.

All original software developed as part of this research is distributed under the MIT License, while the datasets are released under the Creative Commons Attribution 4.0 International License (CC BY 4.0), unless specified otherwise in the respective repositories. These licenses permit broad academic and research use, as well as modification and redistribution, provided appropriate credit is given to the original authors.

In addition to the artifacts created in this project, several external resources were utilized, including pretrained language models (e.g., Llama, OpenAI, DeepSeek, Gemini) and open-source Python libraries such as HuggingFace Transformers and PyTorch. All third-party resources were used strictly in accordance with their respective licenses and intended research purposes, and are appropriately cited in this paper and in the code repositories.

We further note that AI-based tools were used during the preparation of this work. Specifically, AI-guided writing assistants (such as ChatGPT) were employed to refine scientific text, and GitHub Copilot was used to support code development and debugging. The use of these tools was limited to improving clarity and efficiency; all research design, results interpretation, and final manuscript decisions were made by the authors.

The intended use of all released code and data is for research, academic, and educational purposes. Commercial use or deployment in production environments is not permitted without explicit permission or legal review. Any derivatives or extensions of the dataset must comply with the original license terms and the conditions of any incorporated sources. Users are encouraged to consult the individual license files provided in each repository for further details.

\section{Potential Risks}

While this work is primarily intended to advance research in logical reasoning with language models, we recognize several potential risks associated with its development and open release. Enhanced reasoning capabilities in LLMs may be misused, for example, in generating persuasive but misleading arguments, automating manipulation, or circumventing safety mechanisms. The resources and benchmarks we provide, although synthetic and research-focused, could be repurposed for unintended or dual-use applications.

Additionally, while our work does not directly contribute to artificial general intelligence (AGI), we acknowledge broader discussions in the AI community regarding the long-term risks of increasingly capable AI systems. We believe the immediate risks of our work relate to dual-use and misuse as described above, and we encourage responsible use and ongoing monitoring of downstream applications as AI capabilities continue to evolve.

\begin{figure*}[t]
    \centering
    \begin{tcolorbox}[colback=blue!5!white,colframe=blue!75!black,title=\textbf{Synthesis Process and Outputs}, sharp corners]    
    \textbf{Task Specification:}

    (i) \textbf{Language $\mathcal{L}=(\mathcal{V}, \mathcal{G})$}:
    \begin{compactitem}
        \item Vocabulary $\mathcal{V}$: Predicates $\mathcal{P}$ = \{\texttt{is\_red\_train/1, has\_car/2, car\_color/2, car\_len/2}\}; Constants $\mathcal{C}$ = \{\texttt{t1, t2, c1, c2, red, blue, short, long}\}
        \item Grammar $\mathcal{G}$: Restricts predicates to apply to compatible constant types.
    \end{compactitem}
    (ii) \textbf{Configuration $\Theta$}: Rule length $R_{\text{len}}=2$; Problem size $\kappa=(\kappa_{\text{pos}}=1,\ \kappa_{\text{neg}}=1)$
    \hrule
    \vspace{.5em}
    \textbf{Synthesis Steps:}
    
    \noindent\textbf{1. Rule Synthesis:} The \textsc{RuleGenerator} produces a latent ground-truth rule $R^\star$:
    \begin{verbatim}
    is_red_train(T) :- has_car(T, C), car_color(C, red).
    \end{verbatim}
    \vskip -1.2em

    \noindent\textbf{2. Background Synthesis (Loop):}
    
    \textbf{Iteration 1} (finds a positive example):
    \begin{compactitem}
        \item Sample Background ($b_1$): `has\_car(t1, c1). car\_color(c1, red).`
        \item Assign Label: Query $q_1 = \textit{is\_red\_train(t1)}$. Entailment $b_1 \cup R^\star \models q_1$ holds. Result: $(1, q_1)$.
        \item Accept/Reject: $|E^+| < \kappa_{\text{pos}}$, sample is \textbf{accepted}. $B \gets b_1$, $E^+ \gets \{q_1\}$.
    \end{compactitem}
    \textbf{Iteration 2} (finds a negative example):
    \begin{compactitem}
        \item Sample Background ($b_2$): `has\_car(t2, c2). car\_color(c2, blue).`
        \item Assign Label: Query $q_2 = \textit{is\_red\_train(t2)}$. Entailment $b_2 \cup R^\star \models q_2$ fails. Result: $(0, q_2)$.
        \item Accept/Reject: $|E^-| < \kappa_{\text{neg}}$, sample is \textbf{accepted}. $B \gets B \cup b_2$, $E^- \gets \{q_2\}$.
    \end{compactitem}
    The loop terminates as both target sizes are met. The final task is $\mathcal{I} = (B, E^+, E^-)$.
    
    \hrule
    \vspace{.5em}
    \textbf{Final Synthesizer Outputs:}
    
    \noindent\textbf{1. Latent Ground-Truth Rule ($R^\star$):}
    \begin{verbatim}
    is_red_train(T) :- has_car(T, C), car_color(C, red).
    \end{verbatim}
    \vskip -1.2em
    \noindent\textbf{2. Validation Program ($B, E^+, E^-$):}
    \begin{verbatim}
    has_car(t1, c1).
    car_color(c1, red).
    has_car(t2, c2).
    car_color(c2, blue).
    is_red_train(t1).
    \end{verbatim}
    \vskip -1.2em
    \noindent\textbf{3. Instruction Prompt (example formats):}
    
    \textit{(a) Prolog-style Prompt:}
    \begin{verbatim}
    % Given the following background knowledge:
    has_car(t1, c1).
    car_color(c1, red).
    has_car(t2, c2).
    car_color(c2, blue).
    is_red_train(t1).
    % Find a rule "is_red_train(T) :-" that solves the bk.
    \end{verbatim}
    \vskip -1.2em
    \textit{(b) Natural Language Prompt:}
    \begin{verbatim}
    % Given the following background knowledge:
    Train t1 has a car c1. The car c1 is red. 
    Train t2 has a car c2. The car c2 is blue.
    % Find a rule "is_red_train(T) :-" that solves the bk.
    \end{verbatim}
    \end{tcolorbox}
    \vskip -1em
    \caption{Step-by-step example of the automatic ILP task synthesis process in \slr. Given a task specification, comprising a language and a task config, the synthesizer generates a ground-truth rule, samples background knowledge, assigns positive and negative example labels, and produces symbolic (Prolog-style) or natural-language prompts. The figure illustrates all intermediate steps and the final output of the synthesizer.}
    \label{app:fig:synthesis}
\end{figure*}

\begin{table*}[t]
\centering
\setlength{\tabcolsep}{3pt} 
\resizebox{\textwidth}{!}{
\begin{tabular}{clcrrrrrrrr}
\toprule
Org & Model & Reasoning
& LRL
& Syntax
& \multicolumn{4}{c}{Logical-Reasoning Acc. \scriptsize{(\%)$\uparrow$}}
& \multicolumn{2}{c}{Total Compute} \\
\cmidrule(lr){6-9}
\cmidrule(lr){10-11}
& & enabled &
\scriptsize{($\uparrow$0-20)}
& \scriptsize{($\uparrow$\%)}
& Basic & Easy & Medium & Hard
& Tok. \scriptsize{($\downarrow$M)} & Costs \scriptsize{($\downarrow$\$)}\\
\midrule
\openai & o3                           & \reason & \textbf{15.5} &  80 &  99 & \textbf{93} & \textbf{74} & 45 &  4.30 & 207.24 \\
\openai & gpt-5                        & \reason & 15.4 &  98 & \textbf{100} & 90 & 72 & \textbf{46} & 16.40 & 103.13 \\
\openai & gpt-5-mini-high              & \reason & 14.2 &  94 &  99 & 83 & 63 & 38 & 13.12 & 27.98 \\
\openai & o4-mini-high                 & \reason & 12.8 &  88 &  98 & 96 & 40 & 21 &  4.62 & 24.24 \\
\openai & o4-mini                      & \reason & 12.3 &  86 &  93 & 88 & 52 & 13 &  3.98 & 21.43 \\
\openai & gpt-5-mini                   & \reason & 12.0 &  95 &  99 & 82 & 41 & 20 &  4.90 & 11.54 \\
\openai & o1                           & \reason & 11.9 &  68 &  92 & 89 & 41 & 15 &  5.19 & 364.72 \\
\openai & o3-mini                      & \reason & 11.6 &  75 &  97 & 90 & 37 &  7 &  4.73 & 24.71 \\
\qwen   & Qwen3-235B-A22B-Thinking-2507 & \reason & 10.4 &  89 &  82 & 76 & 38 & 12 & 26.32 & 15.53 \\
\openai & o4-mini-low                  & \reason & 10.3 &  91 &  91 & 81 & 25 &  9 &  0.77 & 7.26 \\
\qwen   & Qwen3-235B-A22B-Instruct-2507 & \noreason & 10.3 & 100 &  99 & 76 & 20 & 10 &  5.81 & 4.45 \\
\openai & o1-mini                      & \reason & 10.1 &  95 &  97 & 82 & 20 &  3 &  3.65 & 19.98 \\
\openai & gpt-5-mini-low               & \reason &  9.7 &  88 &  97 & 71 & 20 &  7 &  1.17 & 4.07 \\
\qwen   & Qwen3-Next-80B-A3B-Thinking   & \reason &  9.6 &  96 &  93 & 73 & 18 &  8 & 16.03 & 20.33 \\
\qwen   & Qwen3-32B                    & \reason &  9.0 & 100 &  97 & 64 & 14 &  6 &  7.73 & 1.91 \\
\qwen   & Qwen3-30B-A3B-Thinking-2507   & \reason &  9.0 &  97 &  97 & 67 & 14 &  3 & 17.52 & 4.29 \\
\qwen   & Qwen3-VL-30B-A3B-Thinking     & \reason &  8.9 &  83 &  94 & 70 & 12 &  3 & 18.07 & 11.94 \\
\qwen   & Qwen3-Next-80B-A3B-Instruct   & \noreason &  8.8 & 100 &  93 & 64 & 15 &  5 &  8.68 & 7.68 \\
\deepseek & R1-Llama-70B\footnotemark[2]& \reason &  8.8 &  75 &  98 & 67 &  8 &  4 & 11.61 & 5.33 \\
\google & Gemini-thinking\footnotemark[1] & \reason &  8.6 &  83 &  93 & 65 & 13 &  1 & ---\phantom{0} & ---\phantom{0} \\
\openai & gpt-5-nano                   & \reason &  8.5 &  99 &  97 & 61 & 10 &  2 &  6.17 & 2.81 \\
\qwen   & Qwen3-VL-32B-Thinking         & \reason &  8.4 &  99 &  86 & 66 & 14 &  3 & 19.33 & 23.82 \\
\qwen   & Qwen3-4B-Thinking-2507        & \reason &  8.1 &  99 &  98 & 56 &  8 &  2 & 12.71 & 0.43 \\
\qwen   & Qwen3-30B-A3B-Instruct-2507   & \noreason &  8.0 & 100 &  95 & 57 &  5 &  3 &  7.46 & 2.08 \\
\qwen   & Qwen3-30B-A3B                 & \reason &  7.9 & 100 &  93 & 57 &  7 &  1 &  8.62 & 2.33 \\
\qwen   & Qwen3-VL-8B-Thinking          & \reason &  7.7 &  97 &  87 & 55 &  9 &  2 & 20.77 & 10.44 \\
\qwen   & Qwen3-8B                      & \reason &  7.6 & 100 &  93 & 50 &  7 &  3 & 10.17 & 1.66 \\
\qwen   & Qwen3-VL-4B-Thinking          & \reason &  7.6 &  86 &  91 & 52 &  6 &  2 & 0.00 & 0.00 \\
\qwen   & Qwen3-14B                     & \reason &  7.5 & 100 &  88 & 53 &  8 &  2 &  8.99 & 2.34 \\
\openai & gpt-4.5-prev                  & \noreason &  7.3 & 100 &  94 & 47 &  5 &  1 &  0.37 & 576.40 \\
\qwen   & Qwen3-4B                      & \reason &  6.9 &  99 &  91 & 41 &  5 &  1 &  9.77 & 0.35 \\
\qwen   & Qwen3-4B-Instruct-2507        & \noreason &  6.6 & 100 &  73 & 50 &  6 &  2 &  9.40 & 0.34 \\
\qwen   & Qwen3-235B-A22B               & \noreason &  6.5 & 100 &  87 & 41 &  3 &  0 &  1.05 & 1.88 \\
\openai & gpt-4o                        & \noreason &  6.2 & 100 &  93 & 29 &  2 &  0 &  0.26 & 20.03 \\
\qwen   & Qwen3-32B                     & \noreason &  5.9 & 100 &  89 & 28 &  2 &  0 &  0.63 & 0.49 \\
\qwen   & Qwen3-14B                     & \noreason &  5.6 & 100 &  86 & 25 &  2 &  0 &  0.90 & 0.56 \\
\meta   & Llama 3.3-70B                 & \noreason &  5.6 & 100 &  90 & 22 &  1 &  0 &  0.49 & 0.82 \\
\openai & gpt-4-turbo                   & \noreason &  5.4 & 100 &  89 & 18 &  2 &  0 &  0.41 & 81.30 \\
\qwen   & Qwen3-Coder-30B-A3B-Instruct  & \noreason &  5.3 & 100 &  88 & 17 &  2 &  0 &  3.32 & 1.27 \\
\qwen   & Qwen3-30B-A3B                 & \noreason &  5.2 & 100 &  85 & 19 &  1 &  0 &  1.31 & 0.73 \\
\qwen   & Qwen3-8B                      & \noreason &  4.6 & 100 &  78 & 15 &  0 &  0 &  1.57 & 0.47 \\
\qwen   & Qwen3-4B                      & \noreason &  4.1 & 100 &  70 & 12 &  1 &  0 &  1.47 & 0.15 \\
\meta   & Llama 3.1-8B                  & \noreason &  3.8 &  80 &  70 &  7 &  0 &  0 &  0.20 & 0.14 \\
\qwen   & Qwen3-VL-30B-A3B-Instruct     & \noreason &  3.7 & 100 &  71 &  4 &  0 &  0 &  8.21 & 1.12 \\
\qwen   & Qwen3-VL-32B-Instruct         & \noreason &  3.5 & 100 &  68 &  2 &  0 &  0 &  8.21 & 2.62 \\
\qwen   & Qwen3-VL-8B-Instruct          & \noreason &  3.3 &  99 &  65 &  1 &  0 &  0 &  8.21 & 0.33 \\
\qwen   & Qwen3-1.7B                    & \reason &  1.9 & 100 &  35 &  4 &  0 &  0 &  8.21 & 0.31 \\
\qwen   & Qwen3-VL-2B-Instruct          & \noreason &  1.8 &  58 &  36 &  0 &  0 &  0 & 0.00 & 0.00 \\
\qwen   & Qwen3-0.6B                    & \noreason &  1.5 &  80 &  30 &  0 &  0 &  0 &  4.45 & 0.22 \\
\meta   & Llama 3.2-3B                  & \noreason &  1.0 &  25 &  19 &  1 &  0 &  0 & {0.10} & {0.11} \\
\qwen   & Qwen3-VL-2B-Thinking          & \reason &  0.4 &  87 &   8 &  0 &  0 &  0 & 0.00 & 0.00 \\
\qwen   & Qwen3-1.7B                    & \noreason &  0.3 &  95 &   6 &  1 &  0 &  0 &  1.59 & 0.15 \\
\qwen   & Qwen3-0.6B                    & \reason &  0.2 &  98 &   4 &  0 &  0 &  0 &  5.57 & 0.25 \\
\meta   & Llama 3.2-1B                  & \noreason &  0.0 &  93 &   0 &  0 &  0 &  0 &  0.47 & 0.12 \\
\bottomrule
\end{tabular}
}
\caption{\textbf{Extended Leaderboard.} Logical Reasoning Level (LRL), syntax score, and stage-specific logical reasoning accuracy (basic, easy, medium, hard) and computational costs. Models are ranked by LRL. Models with reasoning mode enabled outperform their base counterparts across reasoning stages, though performance decreases as task complexity rises.}
\label{tab:extended-leaderboard}
\end{table*}

\end{document}